\documentclass[conference]{IEEEtran}
\IEEEoverridecommandlockouts
\usepackage{amsmath,amssymb,amsfonts}
\usepackage{algorithmic}
\usepackage{graphicx}
\usepackage{textcomp}
\usepackage{multirow}
\usepackage{xcolor}
\usepackage{url}
\usepackage{balance}
\usepackage{makecell}
\usepackage{colortbl}
\usepackage{multirow}
\usepackage{tabularx}
\usepackage{booktabs}
\usepackage{bbding}
\usepackage{enumitem}
\usepackage[numbers,sort&compress]{natbib}

\def\BibTeX{{\rm B\kern-.05em{\sc i\kern-.025em b}\kern-.08em
    T\kern-.1667em\lower.7ex\hbox{E}\kern-.125emX}}
\begin{document}

\title{UltraWiki: Ultra-fine-grained Entity Set Expansion with Negative Seed Entities}

\author{
    \IEEEauthorblockN{Yangning Li$^{1,2*}$, Qingsong Lv$^{2*}$, Tianyu Yu$^{2*}$, Yinghui Li$^2$, Xuming Hu$^3$, Wenhao Jiang$^{4\dag}$ \\ Hai-Tao Zheng$^{1,2\dag}$ \textit{Senior Member, IEEE} and Hui Wang$^{1\dag}$}
    \IEEEauthorblockA{}
    \IEEEauthorblockA{$^1$\textnormal{Pengcheng Laboratory}, $^2$\textnormal{Shenzhen International Graduate School, Tsinghua University}}
    \IEEEauthorblockA{$^3$\textnormal{The Hong Kong University of Science and Technology (Guangzhou)} }
    \IEEEauthorblockA{$^4$\textnormal{Guangdong Laboratory of Artificial Intelligence and Digital Economy (SZ)}}
    \IEEEauthorblockA{\textnormal{Email: \{yn-li23, lqs23, liyinghu20\}@mails.tsinghua.edu.cn, zheng.haitao@sz.tsinghua.edu.cn} \\ 
    \textnormal{\{yiranytianyu, xuminghu97, cswhjiang\}@gmail.com, wangh06@pcl.ac.cn}}
    \IEEEauthorblockA{\textnormal{$^*$Equal contribution\quad$^\dag$Corresponding author}}
}

\maketitle

\begin{abstract}
Entity Set Expansion (ESE) aims to identify new entities belonging to the same semantic class as the given set of seed entities. Traditional methods solely relied on positive seed entities to represent the target fine-grained semantic class, rendering them tough to represent ultra-fine-grained semantic classes. Specifically, merely relying on positive seed entities leads to two inherent shortcomings: (i) Ambiguity among ultra-fine-grained semantic classes. (ii) Inability to define ``unwanted'' semantics. Hence, previous ESE methods struggle to address the ultra-fine-grained ESE (Ultra-ESE) task. To solve this issue, we first introduce negative seed entities in the inputs, which jointly describe the ultra-fine-grained semantic class with positive seed entities. Negative seed entities eliminate the semantic ambiguity by providing a contrast between positive and negative attributes. Meanwhile, it provides a straightforward way to express ``unwanted''. 
To assess model performance in Ultra-ESE and facilitate further research, we also constructed UltraWiki, the first large-scale dataset tailored for Ultra-ESE. UltraWiki encompasses 50,973 entities and 394,097 sentences, alongside 236 ultra-fine-grained semantic classes, where each class is represented with 3-5 positive and negative seed entities. Moreover, a retrieval-based framework RetExpan and a generation-based framework GenExpan are proposed to provide powerful baselines for Ultra-ESE. Additionally, we devised two strategies to enhance models' comprehension of ultra-fine-grained entities' semantics: contrastive learning and chain-of-thought reasoning. Extensive experiments confirm the effectiveness of our proposed strategies and also reveal that there remains a large space for improvement in Ultra-ESE. All the codes, dataset, and supplementary notes are available at \url{https://github.com/THUKElab/UltraWiki}.
\end{abstract}


\renewcommand{\thefootnote}{\fnsymbol{footnote}}
\renewcommand{\thefootnote}{\arabic{footnote}}

\section{Introduction}
\begin{figure}
    \centering
    \scalebox{1}{
    \includegraphics[width=1\linewidth]{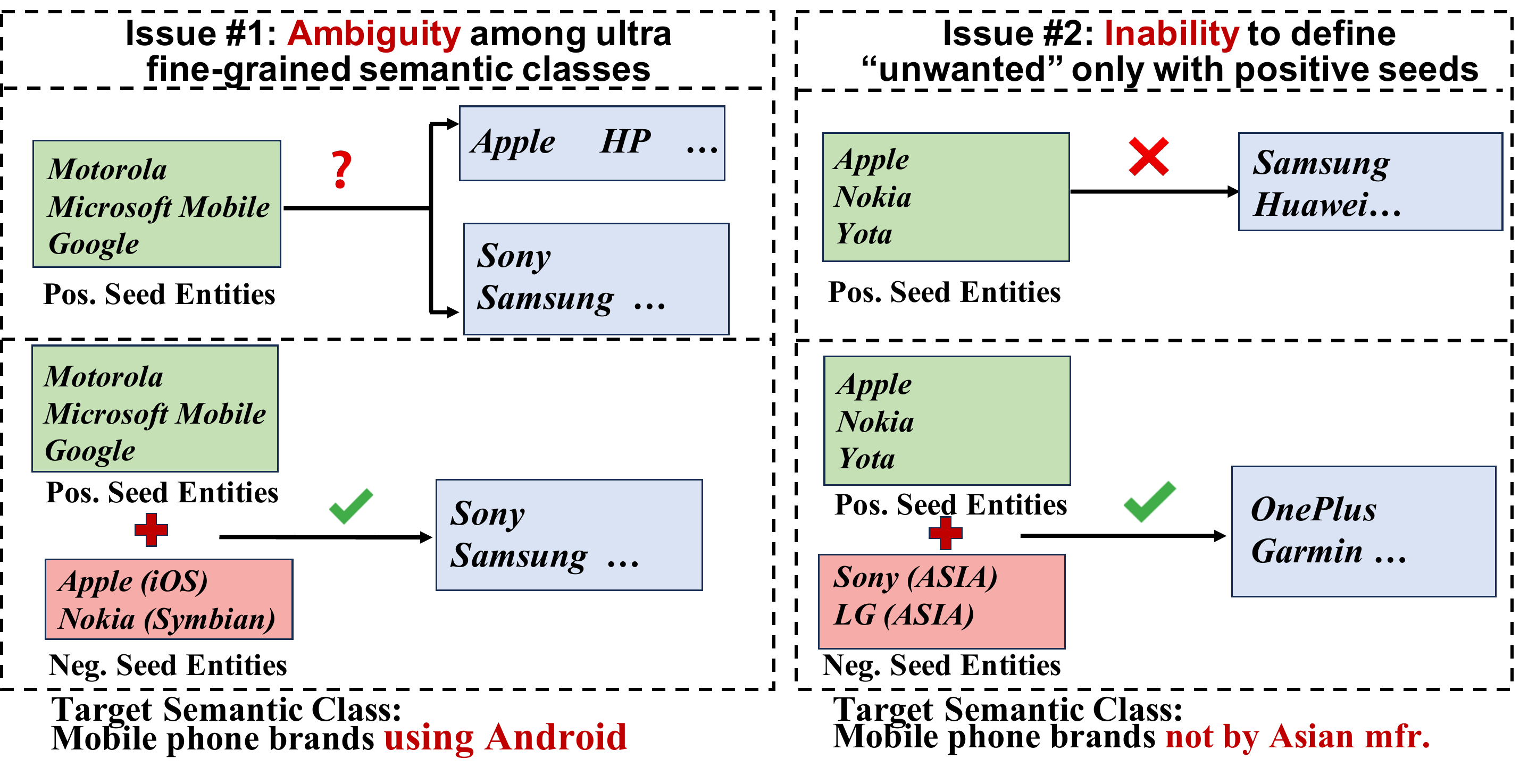}}
    \caption{The figure illustrates how relying only on positive seeds can lead to ambiguity between similar classes (e.g., \texttt{Mobile phone brands using Android} vs. \texttt{American Mobile phone brands}) and the inability to effectively define negative constraints (e.g., \texttt{Mobile phone brands not made by Asian manufacturers}). The introduction of negative seed entities in the input solves these issues.}
    \label{fig:introduction_demo}
\end{figure}
Entity Set Expansion (ESE) is a critical task aiming to expand new entities belonging to the same semantic class as the given set of seed entities \cite{li2023automatic, zhang2020empower,li2022contrastive,li2024mesed}. For example, given the input seed entity set \{\textit{Motorola}, \textit{Xiaomi}, \textit{Nokia}\}, an ESE model will output more new entities (e.g., ``\textit{Huawei}'', ``\textit{Samsung}'', ...) that all belong to the same \texttt{Mobile Phone brands} class as the seed entities.
ESE plays an important role for a wide range of user-tailored applications \cite{wang2019query,seo2021active,whang2016overlapping,pei2021set,kohita2020interactive}, thus demanding a high degree of semantic class granularity. For instance, in the context of recommendation systems \cite{huang2018entity,jacucci2021entity}, a more nuanced ESE method can benefit more precise product recommendations.

Conventionally, existing ESE methods focus on the expansion of ``wanted'' fine-grained semantic classes merely based on positive seed entities, such as \texttt{Car Brand} and the above mentioned \texttt{Phone Brand}. However, current ESE methods struggle to expand \textit{\textbf{ultra}-fine-grained semantic classes} \cite{choi2018ultra, li2023ultra,komarlu2024ontotype}, which involve more specific attribute constraints (e.g., mobile phone brands using Android, Asian mobile phone brands). Since ultra-fine-grained semantic classes from the same concept often exhibit substantial overlap in target entities, rendering them tough to be represented only with a handful of positive seed entities.

Specifically, as shown in Figure \ref{fig:introduction_demo}, merely relying on positive seed entities to represent ultra-fine-grained semantic classes causes two issues: (i) \textbf{Ambiguity among ultra-fine-grained semantic classes.} For instance, with the positive seed entities \{\textit{Motorola}, \textit{Microsoft Mobile}, \textit{Google}\}, it is confusing whether the user intends to extend \texttt{Mobile phone brands using Android} or \texttt{American Mobile phone brands}. Since the positive seed entities alone are insufficient to delineate the distinct class features. (ii) \textbf{Inability to define ``unwanted''.} Positive seed entities alone fail to represent the ``unwanted'' semantic. Considering when users wish to find more \texttt{Mobile phone brands not made by Asian manufacturers} entities, it is impractical to define this semantic class by enumerating brands from other continents in positive seed entities. As the size of the positive seed entities is limited, it's also unreasonable to expect users to specify all potentially desired attribute values.

Due to the inherent shortcomings of the task inputs, prior ESE methods failed to conduct ultra-fine-grained expansion. To address this issue, inspired by exploiting negative feedback to model user intent in recommendation system \cite{zhao2018recommendations,wu2020neural,xie2021deep}, we propose enhancing ESE by incorporating negative seed entities, aiming at ultra-fine-grained ESE (Ultra-ESE). The input negative seed entities belong to the same fine-grained semantic class as the positive seed entities but differs in certain attributes. They jointly represent target semantic class, addressing the semantic granularity issue head-on: First, the contrast between positive and negative seed entities in terms of attributes highlights the user's specific interests and eliminates the ambiguity arising from positive seed entities alone. For the aforementioned example, if negative seed entities are about \texttt{Mobile phone brands not using Android}, it indicates that the user is focused on operating system rather than manufacturer. Second, negative seed entities constrain the expansion space and can readily express ``unwanted'' semantics.

Regrettably, previous ESE dataset \cite{shen2017setexpan,pradhan2013towards} lacked the concept of negative seed entities, let alone ultra-fine-grained semantic classes. To bridge this gap, we constructed UltraWiki, the first large-scale dataset tailored for Ultra-ESE. Derived from Wikipedia, UltraWiki encompasses 10 fine-grained semantic classes, 50,973 entities, and 394,097 sentences. During the curation, we annotated 2-3 attributes for each fine-grained semantic class to further construct ultra-fine-grained semantic classes. Based on the permutations of attributes, 235 ultra-fine-grained semantic classes are constructed, with each semantic class represented with 3-5 positive and negative seed entities. On average, each ultra-fine-grained semantic class contains 23 (236/10-1) hard negative semantic classes, which belong to the same fine-grained semantic class and may exhibit substantial overlap in target entities. Experiments proved that even advancing GPT-4 can not handle it well.

In experiments, we evaluated existing ESE methods on UltraWiki. Furthermore, to comprehensively assess the efficacy of large language models (LLMs) with two different paradigms on the Ultra-ESE task, we designed a retrieval-based framework, RetExpan, with encoder-only LLM BERT~\cite{devlin2018bert} and a generation-based framework, GenExpan, with decoder-only LLM LLaMA~\cite{touvron2023llama}. Meanwhile, we proposed three strategies to enhance models' ability to comprehend ultra-fine-grained semantics of entities: contrastive learning, retrieval augmentation, and chain-of-thought reasoning.

In summary, the main contributions are listed as follows:
\begin{itemize}[fullwidth,itemindent=1.5em]
\item We proposed the more challenging ultra-fine-grained ESE task, and incorporated negative seed entities to represent ultra-fine-grained semantic classes more precisely for the first time.
\item We constructed the first large-scale dataset UltraWiki, tailored for ultra-fine-grained ESE tasks. It encompasses 10 fine-grained semantic classes and 261 ultra-fine-grained semantic classes.
\item We designed both retrieval-based and generation-based frameworks to assess the efficacy of BERT-like and GPT-like LLMs on the UltraWiki dataset. Furthermore, two strategies were proposed for refining the semantic comprehension of ultra-fine-grained entities.
\item Extensive experiments confirmed the effectiveness of our proposed strategies and also reveal significant potential for enhancing ultra-fine-grained semantic comprehension of entities by LLMs.
\end{itemize}

\section{Related Work}

\subsection{Methods of ESE}
Recently, increasing emphasis has been placed on the expansion of fine-grained (e.g., US city) rather than coarse-grained (e.g., Location) semantic classes. ProbExpan \cite{li2022contrastive} employs heuristic methods to mine hard negative entities of target semantic classes, thereby refining the semantic boundaries of fine-grained classes through contrastive learning~\cite{chen2020simple,gao2021simcse}. FGExpan \cite{xiao2023taxonomy} leverages the existing taxonomy to direct BERT in reasoning about more fine-grained names of semantic classes. Besides, MultiExpan \cite{li2024mesed} integrates multimodal pre-trained models to encode visual information (images) associated with entities, which benefits the differentiation of various fine-grained semantic classes. Nevertheless, these methods do not essentially tackle the issue that solely using positive seed entities fails to sufficiently represent ultra-fine-grained semantic classes. In contrast, our research introduces the novel concept of incorporating negative seed entities alongside positive ones to describe an ultra-fine-grained semantic class.

It is worth mentioning that there has been some work to incorporate negative entities \cite{curran2007minimising,jindal2011learning, shi2014probabilistic,li2022contrastive}, though the usage of negative entities in these methods is relatively naive. We point out that the role of the negative entities used in previous work is fundamentally different from ours. Negative entities in previous work are purely used to help determine the boundary of the target set described by positive seed entities. In contrast, negative entities in our model are used to describe the target ultra-fine-grained classes that cannot be characterized by positive seed entities alone.

\subsection{Data Resources of ESE}

Queries in the existing ESE dataset merely consist of positive seed entities. The approach of representing ultra-fine-grained semantic classes solely via positive seed entities inherently faces two major issues: (i) Ambiguity among ultra-fine-grained semantic classes. (ii) Inability to define “unwanted” semantics. Consequently, existing datasets lack ultra-fine-grained semantic classes. For instance, Wiki~\cite{shen2017setexpan} and APR~\cite{shen2017setexpan}, derived from the Wikipedia and Reuters corpora respectively, encompass merely 3 and 8 fine-grained semantic classes. Some named entity recognition datasets are directly used as ESE evaluation benchmarks, they contain semantic classes with coarser granularity. For example, the OntoNotes~\cite{pradhan2013towards} and CoNLL~\cite{sang2003introduction} datasets include semantic classes like organization and location. In contrast, UltraWiki encompasses 261 ultra-fine-grained semantic classes, which surpasses existing ESE datasets in terms of both granularity and quantity of semantic classes.

\section{Task Formulation}
\label{sec:task_form}
We focus on the \textbf{ultra-fine-grained ESE} task, whose input consists of three components: query $S$, candidate entities $V$, and corpus $D$. The query $S = \{ S^{pos} \cup S^{neg}\}$ is a composite set comprising positive seed entities $S^{pos}=\{e^{pos}_1, \ldots, e^{pos}_k\}$ and negative seed entities $S^{neg}=\{e^{neg}_1, \ldots, e^{neg}_k\}$, which belong to the same fine-grained semantic class $c$. The entities in $S^{pos}$ share the same values for positive attribute set $\mathcal{A}^{pos}$: 
\begin{equation}
e.\operatorname{get}(a_i) \equiv k_i, \forall e \in S^{pos}, \forall a_i \in \mathcal{A}^{pos}=\left\{\left(a_i, k_i\right)\right\}_{i=1}^{|\mathcal{A}^{pos}|}
\end{equation}
where $a_i$ and $k_i$ are attribute and value, respectively. We define positive target entities $\mathcal{P}$ as entities that share the same attribute values with positive seed entities in $\mathcal{A}^{pos}$. Likewise, negative target entities in $\mathcal{N}$ share the same attribute values with negative seed entities in $\mathcal{A}^{neg}$. 

\begin{figure}[th]
    \centering
    \includegraphics[width=\linewidth]{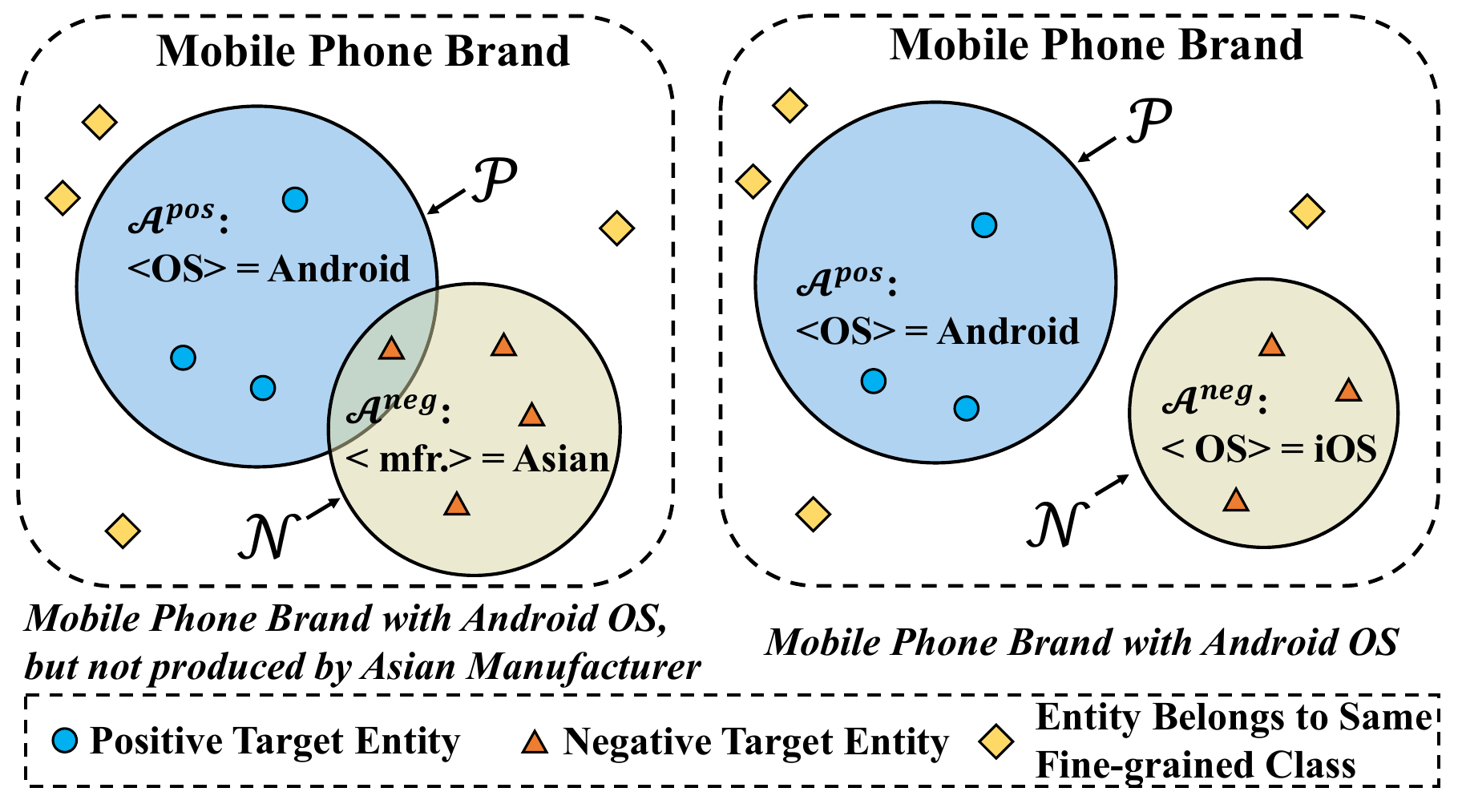}
    \caption{Left target semantic class: mobile phone brand with Android OS, not produced by Asian manufacturer. Right target semantic class: mobile phone brand with Android OS. The target entity set is $\mathcal{P}-\mathcal{N}$.}
    \label{fig:task_form}
\end{figure}

Overall, the target ultra-fine-grained semantic class means class $c$ that holds the same values with positive seed entities in $\mathcal{A}^{pos}$, but different values with negative seed entities in $\mathcal{A}^{neg}$. Examples can be found in Figure \ref{fig:task_form}.

Hence, the ultimate objective of ultra-fine-grained ESE is to identify entities from candidate entities $V$ that belong to the $\mathcal{P}$, while being distinct from $\mathcal{N}$ (e.g., $\mathcal{P}-\mathcal{N}$). In an ideal feature space, entities that share more of the same attribute values should be positioned closer.

\section{UltraWiki Dataset}
\label{sec:dataset}

\begin{figure*}[ht]
    \centering
    \scalebox{0.85}{
    \includegraphics[width=0.85\linewidth]{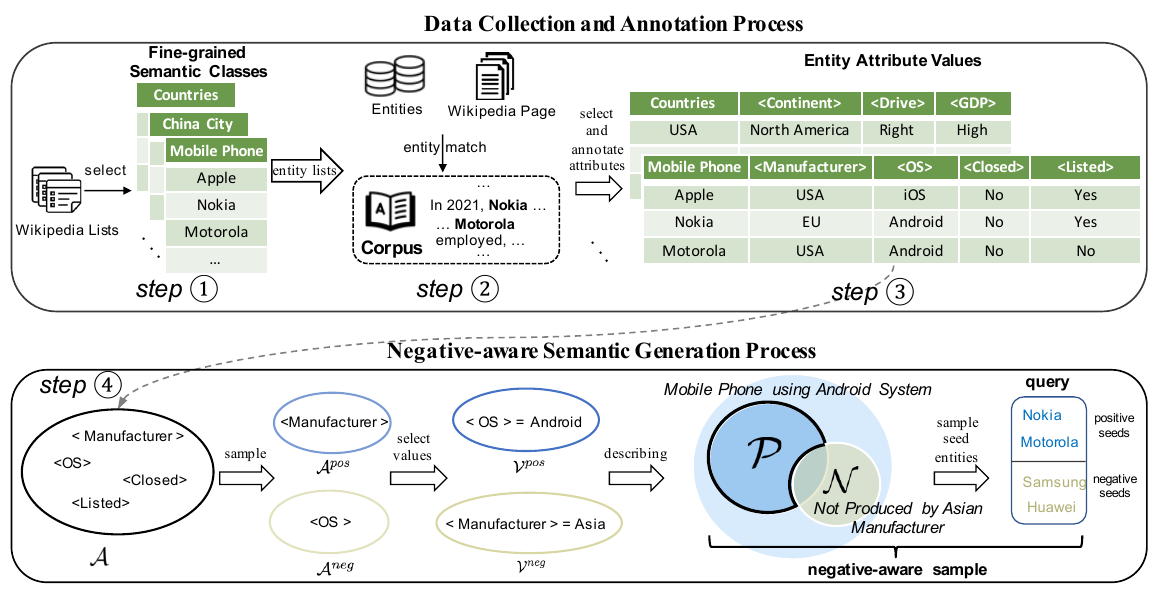}}
    \caption{Illustration of the UltraWiki dataset construction process. The intermediate data obtained at each step is bolded in the figure.}
    \label{fig:data_construct}
\end{figure*}

Considering the lack of ultra-fine-grained semantic classes and negative seed entities in existing ESE datasets, we constructed UltraWiki, the first large-scale dataset tailored for Ultra-ESE. This section outlines the construction process of UltraWiki. Semantic granularity, annotation consistency, and candidate entities' difficulty are fully considered to guarantee the quality of UltraWiki.
\subsection{Dataset Construction}
There are two strategies for constructing UltraWiki. A more direct strategy is to first collect a large number of entities and corresponding context sentences. Then, for each fine-grained semantic class, the human annotator traverses the entire entity vocabulary to identify the corresponding entity and annotates the attribute values, thereby constructing the ultra-fine-grained semantic class. However, despite the availability of numerous entity-centric datasets, this method is labor-intensive and prone to omit entities. Consequently, we chose a more practical strategy. We begin by identifying fine-grained semantic classes along with their corresponding entities, then collecting corresponding sentences and attribute information for each entity in turn. Figure \ref{fig:data_construct} illustrates the entire process.

\noindent\textbf{Step 1. Semantic Classes and Entities Collection.} Wikipedia maintains an extensive catalog of entities associated with fine-grained semantic classes. We identified 10 fine-grained semantic classes and then crawled the corresponding entities. Additionally, a substantial number of entities sampled from Wikipedia pages were incorporated into the candidate entity vocabulary as negative entities.

\noindent\textbf{Step 2. Entity-Labeled Sentences Collection.} A significant volume of text is also crawled from Wikipedia pages, where entities are uniquely identified by hyperlinks. Since the entities obtained in Step 1 also contain hyperlinks, we can readily align entities with sentences containing them, thereby providing abundant contextual information for entities.

\noindent\textbf{Step 3. Entity Attribute Annotation.} For each fine-grained semantic class $C$, we manually select $k$ independent attributes $\mathcal A = \{a_1, a_2, \cdots, a_{k}\}$, which will be used to generate ultra-fine-grained semantic class in next step. Attributes of each class are shown in supplementary notes. Subjective attributes are avoided like good or bad.
We first query Wikidata API for the corresponding attribute value with Python script. For the remaining attributes that could not be automatically annotated, we employed human annotators to manually label them, ensuring each entity was reviewed three times for accuracy and consistency. 

\noindent\textbf{Step 4. Negative-aware Semantic Classes Generation.} We devised an algorithm for automated generation of ultra-fine-grained semantic classes based on annotated attributes. Specifically, for attributes $\mathcal A$ of each fine-grained semantic class $C$, we sample $m$ and $n$ attributes as positive and negative attribute set (i.e., $\mathcal A^{pos}$ and $\mathcal A^{neg}$). Subsequently, we pick a value for all the attributes in $\mathcal A^{pos}$ and $\mathcal A^{neg}$. This process yields a tuple of positive values $\mathcal V^{pos}$ and a tuple of negative values $\mathcal V^{neg}$, which jointly constrain the attributes of the target semantic class to achieve ultra-fine-grained control of the semantic class. 

When $\mathcal A^{pos}$ and $\mathcal A^{neg}$ are the same, the role of $\mathcal A^{neg}$ is to emphasize attributes concerned by user and mitigate ambiguity. Conversely, when they differ, $\mathcal A^{neg}$ serves to convey unwanted semantics. 
As described in the Section \ref{sec:task_form}, positive and negative target entity sets are denoted as $\mathcal{P}$ and $\mathcal{N}$. Both $|\mathcal P|$ and $|\mathcal N|$ are ensured to exceed the minimum entity requirement $n_{thred}=6$.

\begin{figure}[t]
    \centering
    \includegraphics[width=0.9\linewidth]{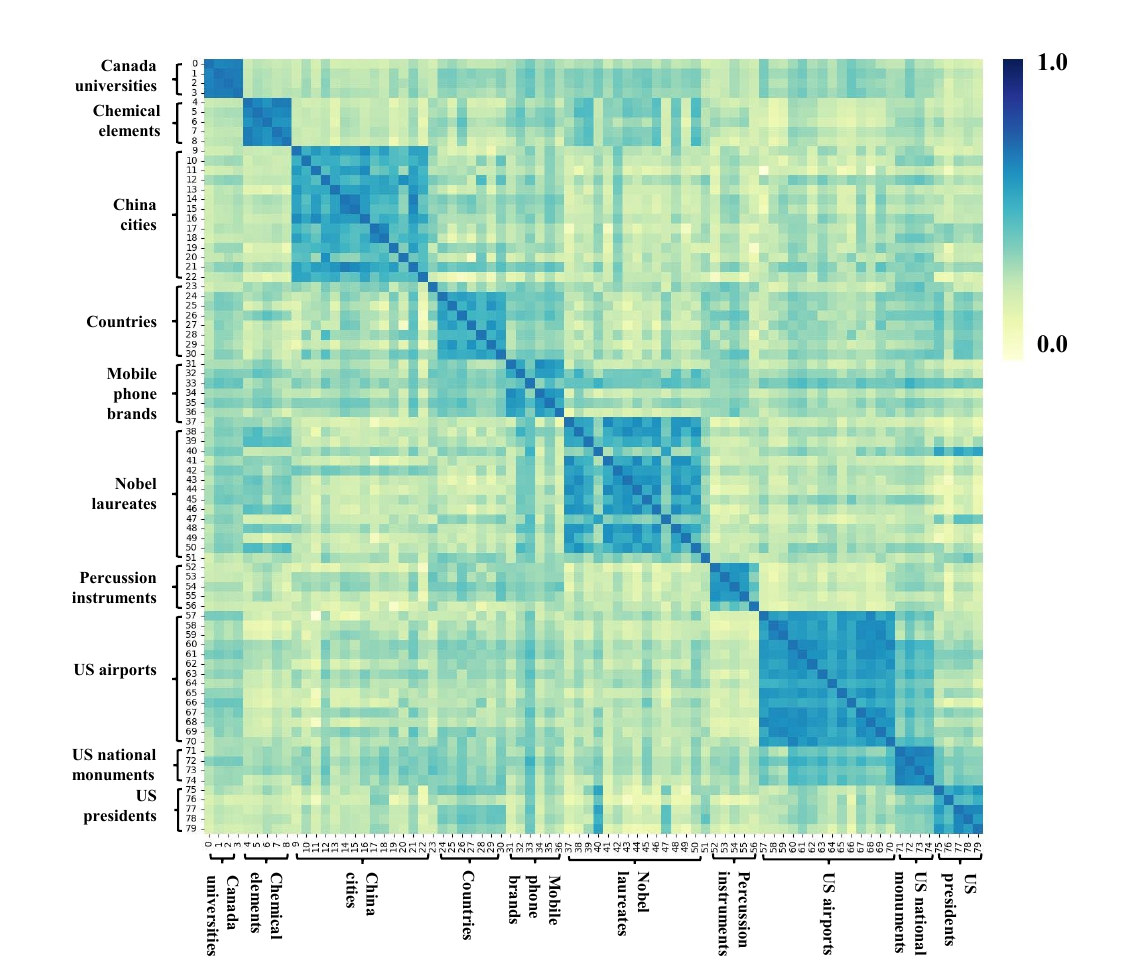}
    \caption{\textcolor{black}{Semantic similarity heatmap of ultra-fine-grained semantic classes. Rows/columns: averaged embeddings of ground-truth positive entities per class; cell colors: pairwise cosine similarities. Classes were proportionally sampled to 80 classes based on total size for clarity.}}
    \label{fig:hitmap}
\end{figure} 
\subsection{Dataset Analysis}
\noindent\textbf{Statistics of UltraWiki.} UltraWiki is the first large-scale dataset featuring ultra-fine-grained semantic classes. It comprises 50,973 entities and 394,097 sentences sourced from Wikipedia. 10 fine-grained semantic classes are determined, that comprehensively cover five major entity types, including Organization, Location, Product, Person, and Miscellaneous. Leveraging the combination of positive and negative attributes, we further derived 261 ultra-fine-grained semantic classes. On average, each ultra-fine-grained semantic class comprises 63 positive target entities ($\mathcal P$) and 60 negative target entities ($\mathcal N$) that are not expected to be expanded. Each ultra-fine-grained semantic class encompasses 3 queries, with 3 to 5 positive and negative seed entities, respectively. The majority of the semantic classes are constrained by one positive and one negative attribute.

\begin{table}[th]
\centering
\caption{Comparison of ESE datasets.}
\scalebox{0.83}{
\begin{tabular}{l|cccccc}
\toprule
 & \textbf{Wiki} & \textbf{APR} & \textbf{CoNLL} & \textbf{ONs} &\textbf{UltraWiki} \\ \midrule
\# Semantic Classes & 8 & 3 & 4 & 8 &261 \\
Semantic granularity & Fine & Fine & Coarse & Coarse &Ultra-Fine \\
\# Queries per Class & 5 & 5 & 1 & 1  & 3 \\
\# Pos Seeds per Query & 3 & 3 & 10 & 10 &3-5\\
\# Neg Seeds per Query & N/A & N/A & N/A & N/A &3-5\\
\# Candidate Entities & 33K & 76K & 6K & 20K &51K \\
\# Sentences of Corpus & 973K & 1043K & 21K & 144K &394K\\
Entity Attribution &\XSolidBrush  &\XSolidBrush&\XSolidBrush&\XSolidBrush&\Checkmark \\
\bottomrule
\end{tabular}}
\label{tab:datastat}
\end{table}

\noindent\textbf{Quality of UltraWiki.} The fine-grained semantic classes, entity corpus, and partial attributes of UltraWiki are automatically crawled from Wikipedia and Wikidata. Both sources are high-quality knowledge bases, curated by numerous domain experts, thereby firmly guaranteeing the quality of UltraWiki. For attributes that require manual annotation, we ensure that each value is annotated by three annotators. Eventually, the inter-annotator agreement measured by Fleiss's Kappa \cite{fleiss1971measuring} reaches 0.90, indicating satisfying consistency and accuracy of the annotation.

\noindent\textbf{Difficulty of UltraWiki.} Ultra-ESE is an inherently challenging task. The challenge arises from the substantial entity overlap among ultra-fine-grained semantic classes (approximately 99\% of ultra-fine-grained semantic classes intersect), which requires the model to comprehend the ultra-fine-grained distinctions of entities across varied contexts. Even advanced models like GPT-4 struggle to address this challenge (refer to the results in the section of experiments). 

Additionally, our selected semantic classes include a subset of long-tail entities, such as lesser-known Chinese cities, which typically lack extensive contextual knowledge. Simultaneously, employing the BM25-based search, we incorporated entities highly similar to the target entities as hard negative entities in the candidate entity vocabulary. The inclusion of these entity types contributes to the difficulty of UltraWiki. 
\textcolor{black}{The visualization of semantic similarity in Figure \ref{fig:hitmap} demonstrates that each semantic class in UltraWiki exhibits remarkably high intra-class similarity, where each row/column represents the average embedding of all ground-truth positive entities within a specific ultra-fine-grained semantic class.}

\begin{figure*}[t]
    \centering
    \scalebox{0.8}{
    \includegraphics[width=0.9\linewidth]{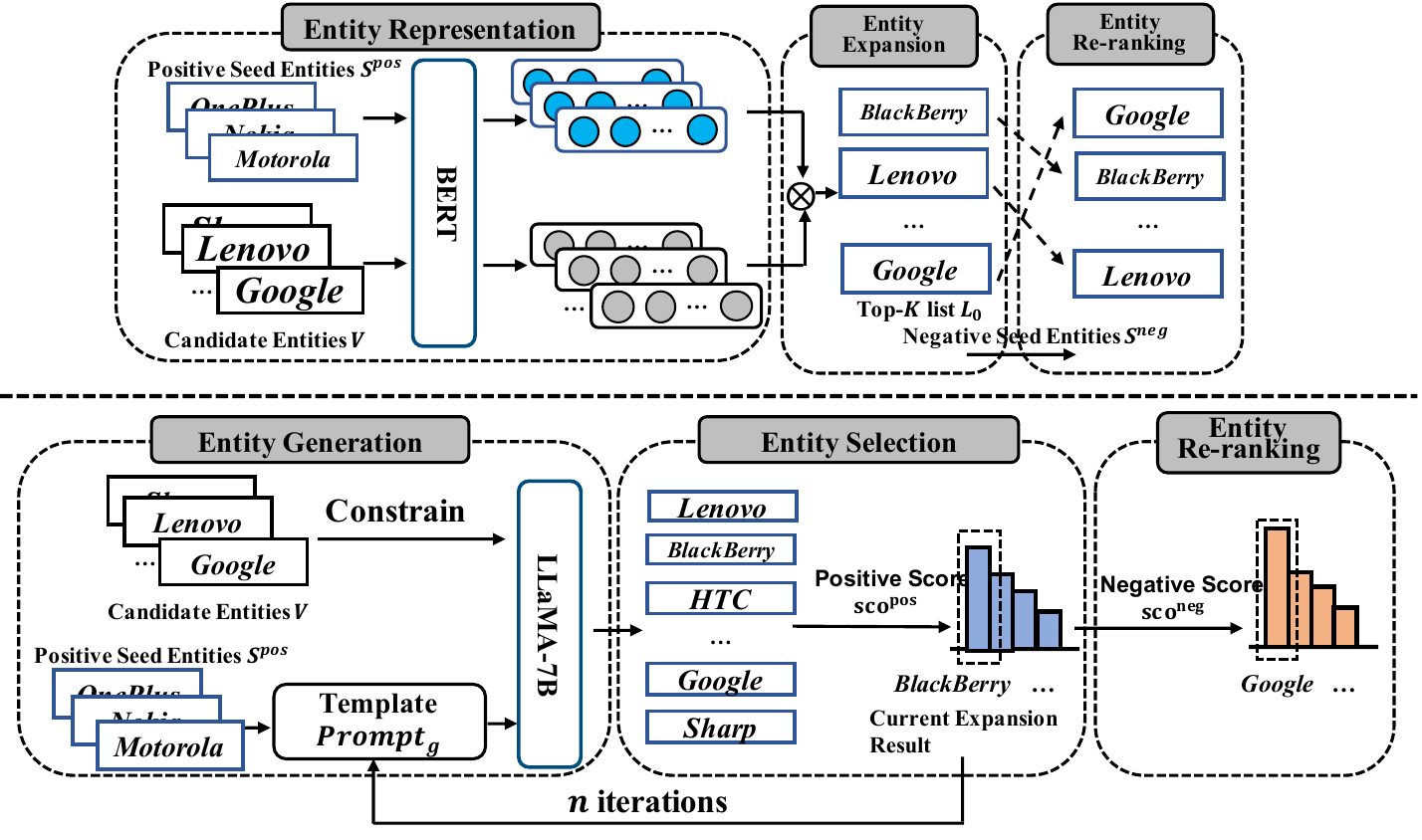}}
    \caption{Upper: overall frameworks of our RetExpan. Below: overall frameworks of our GenExpan.}
    \label{fig:framework}
    \vspace{-0.5cm}
\end{figure*}
\section{Methods}
This section presents our proposed retrieval-based and generation-based frameworks to leverage both positive and negative seed entities, namely RetExpan and GenExpan. Here, we re-emphasize that the inputs of the two frameworks comprise both positive and negative seed entities, which belong to the same fine-grained semantic class and solely differ in their ultra-fine-grained attributes.

Most prior ESE models are built upon retrieval-based frameworks, e.g., CGExpan and ProbExpan. The retrieval-based framework first encodes entities as features in low-dimensional semantic space, and then measures the probability that a candidate entity belongs to the target semantic class via entity feature similarity, which is naturally more suitable for retrieval tasks like ESE. With the advent of generative LLMs, we note their significant semantic comprehension and reasoning skills, which are developed through pre-training on large-scale corpora. These models can eliminate the intermediate step of entity embedding, allowing for a more streamlined, end-to-end Entity Set Expansion ESE process. Therefore, we also propose a generation-based framework to explore the potential of generative LLMs. Moreover, for both frameworks, enhancement strategies are crafted to enhance the models' capacity for ultra-fine-grained semantic comprehension with negative seed entities. The proposed RetExpan and GenExpan provide comprehensive and strong baselines for Ultra-ESE.

\subsection{Retrieval-based Framework: RetExpan}
\subsubsection{Overall Framework}
 RetExpan is structured into three steps: entity representation, entity expansion, and entity re-ranking. In the first stage, we design the entity encoder that extracts contextual features of entities in sentences. An entity is represented as the mean of the feature vectors derived from all sentences containing it. The entity prediction task is introduced to refine the entity representation. During the second stage, a preliminary list of target entities is acquired based on their similarity to the positive seed entities. Notably, negative seed entities are excluded in this process to ensure the recall of all entities satisfying fine-grained semantic classes. In the third stage, negative seed entities are also employed to re-rank the entity list obtained in the previous phase, reducing the ranking of entities aligning with negative attribute values.

\noindent\textbf{Entity Representation.} 
The goal of entity encoder is to extract contextual features of entities from textual data. Initially, we replace entity mentions within a sentence with [MASK] tokens to construct the input for the encoder. Given a contextual sentence $T$ with masked entity mentions, we utilize the $\text{BERT}_{\text{BASE}}$, which comprises a 12-layer Transformer~\cite{vaswani2017attention}, to extract contextual embeddings:
\begin{equation}
    H = \{{h}_1, {h}_2, ..., {h}_{L}\}=\text{BERT}_{\text{BASE}}(T)
\end{equation}
where $L$ is the length of the tokenized sentences. Ultimately, the contextual feature $\mathbf{h}$ of an entity is computed as the average of the contextual embedding mask $h_{[\text{MASK}]}$ at the mask position across all sentences containing it.

Inspired by ProbExpan, the entity prediction task is introduced to refine the entity representation. Concretely, a classification header $\mathbf{f}$ is appended to the entity encoder. After obtaining the hidden state $h_{[\text{MASK}]}$ at [MASK] position, it is transformed into the probability distribution of the masked entity among the candidate entities via MLP and SoftMax:
\begin{equation}
\hat{y}=\mathbf{f}\left(h_{[\text{MASK}]}\right)=\operatorname{Softmax}\left(\operatorname{MLP}\left(h_{[\text{MASK}]}\right)\right), \hat{y} \in \mathbb{R}^{V_e}
\end{equation}
where $V_e$ is the size of candidate entities. Then, cross-entropy loss with label smoothing~\cite{szegedy2016rethinking} is applied to learn the latent semantics of entities:
\begin{equation}\label{eq:mep}
\begin{split}
    \mathcal{L}_{mask} = -\frac{1}{N} \sum_i^{N} \sum_{j}^{V_e} y_{i}[j]\cdot(1-\eta)\cdot log(\hat{y}_{i}[j]) \\ +(1-y_{i}[j])\cdot \eta \cdot log(1-\hat{y}_{i}[j])
\end{split}
\end{equation}
in which ground-truth $y$ is a one-hot vector and $N$ is the batch size. Smoothing factor $\eta$ mitigates over-penalization for entities that exhibit similar semantics to the ground-truth entity.

\noindent\textbf{Entity Expansion.} We acquire the preliminary entity list comprising entities that belong to the same fine-grained semantic class as the positive seed entities ${S}^{pos}$. For each candidate entity $e$, we define its positive similarity score $sco^{pos}$ as follows:
\begin{equation}
\label{eq:retrive_score}
sco^{pos}(e)=\frac{1}{|{S}^{pos}|} \sum_{e'\in {S}^{pos}} \mathrm{cos}\_\mathrm{sim}\left( h(e), h(e')\right)
\end{equation}
We keep top-$K$ entities with the highest $sco^{pos}$ as the initial expansion list $L_0$.

\noindent\textbf{Entity Re-ranking.} This step further considers negative seed entities, excluding entities that share semantics with them (i.e., possess the same value on negative attribute). Analogous to Equation \ref{eq:retrive_score}, we define the negative similarity score ${sco}^{neg}$ between the candidate entity and negative seed entities. However, directly re-ranking $L_0$ in ascending order utilizing $sco^{neg}$ introduces a significant number of noisy entities that do not belong to the same fine-grained semantic class as the seed entities. These irrelevant entities, characterized by low similarity scores with the negative seed entities, may be erroneously assigned over-high ranking.

To address this issue, we propose a simple yet effective strategy: segmented re-ranking. This approach divides $L_0$ equally into $\left\lceil\frac{\left|L_0\right|}{l}\right\rceil$ segments and then conducts a descending re-ranking using $sco^{neg}$ for each segment individually. Hyper-parameter $l$ represents the length of each segment. This strategy facilitates a local fine ranking of $L_0$ and prevents the assignment of over-high rankings to noisy entities with quite low $sco^{neg}$ values.

\subsubsection{Enhancement Strategy 1: Ultra-fine-grained Contrastive Learning} 
To further enhance the ultra-fine-grained semantic comprehension capability of RetExpan, we devised attribute-aware contrastive learning. Conventional contrastive learning \cite{robinson2021contrastive} facilitates clearer semantic boundaries for fine-grained semantic classes by drawing the representation of the same semantic class entities closer and the representation of different semantic class entities further apart. Although proven to be effective, basic contrastive learning is not precise enough for Ultra-ESE as it neglects entity differences inside the same semantic class. For instance, ``Samsung'' and ``Motorola'' may serve as positive pairs in terms of the attribute ``Operating System'' but as negative pairs for the attribute ``Manufacturer''. Ideally, entities with more attributes in common should exhibit closer distance in feature space. To this end, we construct ultra-fine-grained training data based on positive and negative seed entities to implicitly incorporate attribute-based differences into entity representations.

\noindent\textbf{Ultra-fine-grained Training Data.} We prompt GPT-4\footnote{All specific prompts used for querying GPT-4 can be found in supplementary notes. We use GPT-4-1106.} to deduce potential positive (negative) attributes and return T entities from the initial list $L_0$ that are most similar to the given positive (negative) seed entities to form the entity list $L_{pos}$ ($L_{neg}$).  Entities in $L_{pos}$ and $L_{neg}$ are roughly considered to belong to the positive (target) and negative ultra-fine-grained semantic class $\mathcal{P}$ and $\mathcal{N}$, respectively. Hence, to distinguish entities with different attribute values under the same fine-grained semantic class, entities in $L_{pos}$ and $L_{neg}$ should be pulled away from each other. Specifically, we construct the contrastive training pairs as follows:
\begin{equation}
\small
\begin{split}
\mathbb{P}_{pos}=\{(x,x')|e(x)\in L_{pos}, e(x')\in L_{pos}\} \cup \\
    \{(x,x')|e(x)\in L_{neg}, e(x')\in L_{neg}\} \cup \{(x,x')|e(x)=e(x')\},
\end{split}
\end{equation}
\begin{equation}
\small
\begin{split}
\mathbb{P}_{neg}=\{(x,y)|e(x)\in L_{pos},e(y)\in L_{neg}\}\cup\\
    \{(x,z)|e(x)\in L_{pos}\cup L_{neg},e(z)\in\overline{L_0}\},
\end{split}
\end{equation}
where $e(x)$ indicates the entity of the training sample $x$. $\overline{L_0}$ denotes entities from other fine-grained semantic classes, which form normal negative pairs with entities from $L_{pos}$ and $L_{neg}$. $\overline{L_0}$ is necessary to ensure that from the fine-grained semantic level, entities in $L_{pos}$ and $L_{neg}$ remain similar in the semantic space. Moreover, during the training process, we append the corresponding positive and negative seed entities after each training sample (i.e., sentence) to implicitly specify the corresponding ultra-fine-grained semantics, avoiding positive-negative conflicts for the same entity pair.

\noindent\textbf{Contrastive Loss.} We choose the classical InfoNCE loss \cite{oord2018representation} as the contrastive loss. Contrastive learning is conducted in a new hypersphere space to prevent semantic collapse, which is transformed by another MLP-based mapping head $\mathbf{f_{cl}}$ and $l$-2 normalization.

\subsection{Generation-based Framework: GenExpan}
\subsubsection{Overall Framework}
To explore the potential of emerging generative LLMs, we devised the generative framework GenExpan. Unlike RetExpan which relies on intermediate semantic features of entities, GenExpan directly infers based on the given preceding text and knowledge stored in LLMs themselves, thus is more efficient. The proposed GenExpan consists of three phases: entity generation, entity selection, and entity re-ranking. In the first phase, GenExpan generates a new set of entities based on positive seed entities and current expanded entities. A constrained decoding strategy is designed to ensure that the generated entities are included in the candidate entities. Subsequently, the entity selection phase filters $K$ entities to be added to the current expansion, based on the similarity scores between the generated entities and positive seed entities. The first two steps are executed iteratively for several rounds. Finally, akin to RetExpan, GenExpan re-ranks the previous expansion results to lower the ranking of entities identical to the negative seed entities in terms of negative attributes. GenExpan utilizes LLaMA-7B \cite{touvron2023llama} as the backbone. Additionally, we continually pre-trained LLaMA-7B with the given corpus ${D}$ to fully leverage entity semantic information. 

\begin{figure}
    \centering
    \includegraphics[width=0.9\linewidth]{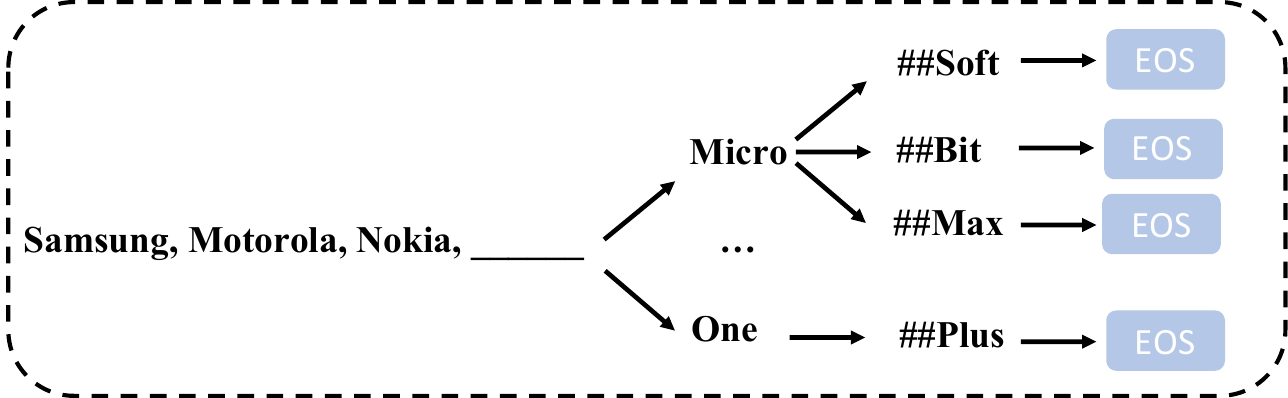}
    \caption{Prefix tree-based constrained decoding.}
    \label{fig:prefix_tree}
    \vspace{-0.5cm}
\end{figure}

\noindent\textbf{Entity Generation.} We craft a simple prompt $\text{Prompt}_\text{g}$ with the given entity set to guide the model toward generating entities semantically similar to the target semantic class. The full prompt can be found in supplementary notes. In the first round of generation, we randomly select 3 entities from positive seed entities $S^{pos}$ as the input entity set. In subsequent rounds, to maintain diversity while ensuring the semantic does not deviate from original positive seed entities, 2 entities and 1 entity are randomly sampled from $S^{pos}$ and current expansion results, respectively.

Vanilla decoding strategies like beam search~\cite{sutskever2014sequence} might uncontrollably generate entities that are not included in the candidate entities. To address this issue, a prefix tree structure is constructed to constrain the decoding process. In this structure, the root node represents the beginning token of an entity, and each path from the root to a leaf node represents a complete candidate entity. During decoding, the process must follow a specific path from root to leaf at one time. In other words, the child nodes of each node represent the tokens allowed for subsequent generation. By employing this prefix-constrained beam search, we ensure the generated entities remain valid.

\noindent\textbf{Entity Selection.} For each generated entity $e$, we calculate the positive similarity score $sco^{pos}$ between it and positive seed entities $S^{pos}$, which is formulated as follows:
\begin{equation}
sco^{pos}(e)=\frac{1}{|S^{pos}|} \sum_{e'\in S^{pos}}\sqrt[|e'|]{P(e'|f(e))},
\end{equation}
in which $f(e)$ is the template ``\{e\} is similar to''. $P(e'|f(e))$ represents the conditional probability of LLM generating $e'$ when given $f(e)$. We use geometric mean to balance the various token length $|e'|$ of entities. Top-$p$ entities with the highest positive similarity scores are incorporated into the current expansion result.

\noindent\textbf{Entity Re-ranking.} The process of reranking in GenExpan is identical to that in RetExpan except for the calculation of $sco^{neg}$.

\subsubsection{Enhancement Strategy 2: Chain-of-thought Reasoning}
Inspired by \cite{wei2022chain}, we enhance the reasoning capability of LLMs for Ultra-ESE through a series of intermediate reasoning steps. Basically, the positive seed entities $S^{pos}$ in GenExpan are directly utilized to construct $\text{Prompt}_\text{g}$ for generating new entities. However, in chain-of-thought reasoning, we prompt the LLM to initially generate potential fine-grained class names of the positive seed entities, along with the positive attributes that share the same value. This information is then integrated into $\text{Prompt}_\text{g}$ to guide subsequent entity generation. The chain-of-thought reasoning stimulates the reasoning capability of LLM and benefits it for perceiving the semantic classes of seed entities and the concerned attributes.

\section{Experiments}
\subsection{Experiment Setup}
\noindent\textbf{Compared Methods.} We compare with three categories of models. The initial category encompasses traditional statistical probability-based methods, notably \textbf{SetExpan}~\cite{shen2017setexpan} and \textbf{CasE}~\cite{yu2019corpus}. The second category comprises methods based on the pre-trained language model BERT, including \textbf{CGExpan} and \textbf{ProbExpan}, with ProbExpan representing the prior state-of-the-art method. Lastly, the third category involves methods based on the generative LLM, \textbf{GPT-4}. We devised prompt templates incorporating both positive and negative seed entities to guide the model to generate target entities.

\noindent\textbf{Evaluation Metrics.} The primary objective of ultra-fine-grained ESE is to expand the entity list in descending order based on their similarity with positive seed entities. Following prior research, PosMAP@$K$ and PosP@$K$ are employed as evaluation metrics. PosMAP@$K$ is computed as follows:
\begin{equation}
\mbox{PosMAP@}K=\frac{1}{|Q|} \sum_{q \in Q} \mbox{PosAP}_K\left(L_{q}, \mathcal P\right),
\end{equation}
where $Q$ stands for the set of all queries, $ \mbox{PosAP}_K\left(L_{q}, \mathcal P\right)$ denotes the average precision at position $K$ with the ranked list $L_{q}$ and ground-truth list $\mathcal{P}$. PosP@$K$ is the precision of the
top-$K$ entities. 

Notably, another crucial objective of ultra-fine-grain ESE is to prevent the intrusion of negative entities in the candidate entity list that satisfy the negative attribute (entities in $\mathcal{N}$). 
Hence, we symmetrically define NegMAP@$K$ and NegP@$K$, whereby computing them only requires substituting $\mathcal{P}$ with $\mathcal{N}$. A well-built model should also keep negative metrics as low as possible. The combined metrics $\mbox{CombMAP@}K=\left(\mbox{PosMAP@}K+100-\mbox{NegMAP@}K\right)/2$, and $\mbox{CombP@}K=\left(\mbox{PosP@}K+100-\mbox{NegP@}K\right)/2$ are also calculated to comprehensively reflect model capacities. These combined metrics are normalized to range from 0 to 1, with higher values indicating better performance. In summary, our evaluation metric can be expressed as $xy@K$, where $x$=\{Pos, Neg, Comb\}, y=\{MAP, P\}, $K$=\{10, 20, 50, 100\}.

\begin{table*}[]
\centering
\caption{We report the MAP and P scores corresponding to Pos/Neg/Comb on UltraWiki. The highest score is bolded.}
\scalebox{0.85}{
\begin{tabular}{lllccccccccc}
\toprule
\multirow{2}{*}{\textbf{Metric Type}} & \multirow{2}{*}{\textbf{Method Type}}                                                         & \multirow{2}{*}{\textbf{Method}} & \multicolumn{4}{c}{\textbf{MAP}}       & \multicolumn{4}{c}{\textbf{P}}         & \multirow{2}{*}{\textbf{Avg}} \\ \cmidrule{4-11}
                             &                                                                                      &                         & \textbf{@10}   & \textbf{@20}   & \textbf{@50}   & \textbf{@100}  & \textbf{@10}   & \textbf{@20}   & \textbf{@50}   & \textbf{@100}  &                      \\ \midrule
\multirow{11}{*}{Pos $\uparrow$}        & Probability Based         & SetExpan                & 13.41 & 11.83 & 10.66 & 10.79 & 20.10 & 19.97 & 21.88 & 26.39 & 16.88                \\
                             &                                                                                      & CaSE                    & 16.72 & 13.74 & 11.60 & 10.91 & 24.58 & 23.28 & 26.86 & 30.53 & 19.78                \\ \cmidrule{2-12} 
                             & Retrieval Based           & CGExpan                 & 21.64 & 19.72 & 19.11 & 20.22 & 30.61 & 31.24 & 38.39 & 50.03 & 28.87                \\
                             &                                                                                      & ProbExpan               & 21.86 & 22.11 & 22.80 & 23.89 & 38.08 & 39.41 & 47.02 & 62.71 & 34.74                \\ \cmidrule{2-12} 
                             & Generation Based                           & GPT4                    & 37.20 & 35.37 & 35.49 & 35.59 & 47.12 & 48.87 & 55.31 & 62.22 & 44.65                \\ \cmidrule{2-12} 
                             & \multirow{2}{*}{\begin{tabular}[c]{@{}l@{}}Retrieval Based\\ (Ours)\end{tabular}}  & RetExpan                & 41.73 & 39.53 & 38.55 & 39.91 & 54.58 & 58.03 & 66.76 & 77.23 & 52.04                \\
                             &                                                                                      & RetExpan + Contrast              & 47.45 & 44.68 & 43.63 & \textbf{44.20} & 59.83 & 62.02 & \textbf{69.36} & \textbf{77.92} & \textbf{56.14}                               \\ \cmidrule{2-12} 
                             & \multirow{2}{*}{\begin{tabular}[c]{@{}l@{}}Generation Based\\ (Ours)\end{tabular}} & GenExpan                & 46.79 & 45.00 & 42.89 & 40.80 & 59.77 & 62.15 & 66.26 & 66.57 & 53.78                \\
                             &                                                                                      & GenExpan + CoT                   & \textbf{50.39} & \textbf{47.80} & \textbf{43.67} & 40.06 & \textbf{62.74} & \textbf{64.45} & 64.06 & 60.38 & 54.19                               \\ \midrule
\multirow{11}{*}{Neg $\downarrow$}        & Probability Based         & SetExpan                & 4.06  & 3.77  & 3.71  & 4.20  & 7.66  & 8.10  & 10.92 & 17.44 & 7.48                 \\
                             &                                                                                      & CaSE                    & 5.33  & 4.32  & 3.63  & 3.50  & 10.22 & 10.10 & 12.79 & 15.96 & 8.23                 \\ \cmidrule{2-12} 
                             & Retrieval Based           & CGExpan                 & 6.15  & 6.54  & 8.03  & 9.96  & 12.29 & 16.37 & 27.38 & 41.72 & 16.06                \\
                             &                                                                                      & ProbExpan               & 6.72  & 8.16  & 10.85 & 13.47 & 15.12 & 19.92 & 34.51 & 56.48 & 20.65                \\ \cmidrule{2-12} 
                             & \begin{tabular}[c]{@{}l@{}}Generation Based\end{tabular}                           & GPT4                    & 6.04  & 6.61  & 8.03  & 8.35  & 10.40 & 15.06 & 24.57 & 33.63 & 14.09                \\ \cmidrule{2-12} 
                             & \multirow{2}{*}{\begin{tabular}[c]{@{}l@{}}Retrieval Based\\ (Ours)\end{tabular}}  & RetExpan                & 8.77  & 9.04  & 10.65 & 13.29 & 16.44 & 21.04 & 34.78 & 56.54 & 21.32                \\
                             &                                                                                      & RetExpan + Contrast              & 8.02  & 8.98  & 10.89 & 13.05 & 14.83 & 21.23 & 35.47 & 55.12 & 20.95                \\ \cmidrule{2-12} 
                             & \multirow{2}{*}{\begin{tabular}[c]{@{}l@{}}Generation Based\\ (Ours)\end{tabular}} & GenExpan                & 7.25  & 8.28  & 8.72  & 8.01  & 15.21 & 21.31 & 27.33 & 28.58 & 15.59                \\
                             &                                                                                      & GenExpan + CoT                   & 7.79  & 9.29  & 8.15  & 6.89  & 15.97 & 22.66 & 23.52 & 21.90 & 14.52                \\ \midrule
\multirow{11}{*}{Comb $\uparrow$}       & Probability Based         & SetExpan                & 54.67 & 54.03 & 53.48 & 53.30 & 56.22 & 55.93 & 55.48 & 54.48 & 54.70                \\
                             &                                                                                      & CaSE                    & 55.69 & 54.71 & 53.99 & 53.70 & 57.18 & 56.59 & 57.03 & 57.28 & 55.77                \\ \cmidrule{2-12} 
                             & Retrieval Based           & CGExpan                 & 57.75 & 56.59 & 55.54 & 55.13 & 59.16 & 57.44 & 55.50 & 54.15 & 56.41                \\
                             &                                                                                      & ProbExpan               & 57.57 & 56.98 & 55.97 & 55.21 & 61.48 & 59.75 & 56.25 & 53.12 & 57.04                \\ \cmidrule{2-12} 
                             & \begin{tabular}[c]{@{}l@{}}Generation Based\end{tabular}                           & GPT4                    & 65.58 & 64.38 & 63.73 & 63.62 & 68.36 & 66.90 & 65.37 & 64.29 & 65.28                \\ \cmidrule{2-12} 
                             & \multirow{2}{*}{\begin{tabular}[c]{@{}l@{}}Retrieval Based\\ (Ours)\end{tabular}}  & RetExpan                & 66.48 & 65.25 & 63.95 & 63.31 & 69.07 & 68.50 & 65.99 & 60.34 & 65.36                \\
                             &                                                                                      & RetExpan + Contrast              & 69.72 & 67.85 & 66.37 & 65.57 & 72.50 & 70.39 & 66.95 & 61.40 & 67.59                \\ \cmidrule{2-12} 
                             & \multirow{2}{*}{\begin{tabular}[c]{@{}l@{}}Generation Based\\ (Ours)\end{tabular}} & GenExpan                & 69.77 & 68.36 & 67.08 & 66.40 & 72.28 & 70.42 & 69.47 & 68.99 & 69.10                \\
                             &                                                                                      & GenExpan + CoT                   & \textbf{71.30} & \textbf{69.25} & \textbf{67.76} & \textbf{66.58} & \textbf{73.39} & \textbf{70.90} & \textbf{70.27} & \textbf{69.24} & \textbf{69.84}                       \\ \bottomrule    
\end{tabular}}
\label{tab:main2}
\end{table*}

\subsection{Main Experiments}

The results of the main experiment are presented in Table \ref{tab:main2},  from which we can observe that:

(1) Regarding the average values of the comprehensive metrics CombMAP and CombP, our proposed GenExpan achieves optimal and sub-optimal performances with the incorporation of the chain-of-thought reasoning strategy and the entity-based retrieval augmentation strategy, respectively. Although the backbone model of GenExpan, LLaMA, has only 7b parameters, it surpasses the GPT-4, which comprises more than 1T parameters (over 200 times larger than the GenExpan). We attribute this to two factors. Firstly, our LLaMA undergoes further pretraining on the provided corpus focusing on entities, thereby reinforcing the model's ultra-fine-grained semantic understanding of entities. Secondly, GenExpan ensures that the generated entities belong to the candidate entities by constraining the decoding process through the prefix tree. This mitigates the negative impact of introducing irrelevant entities on PosMAP and PosP. It can also be seen from the results that GenExpan outperforms GPT-4 mainly in the Pos class metrics. This constraint strategy is particularly valuable for ESE in user-customized scenarios. It ensures that the entities output by GenExpan are within user's focus domain.

(2) Our proposed RetExpan model also consistently outperforms the leading retrieval-based model, ProbExpan, and the advanced generation-based model, GPT-4, as measured by the Comb metrics. RetExpan and ProbExpan are similar in their overall frameworks but have a large performance gap. The primary reason for this difference is that RetExpan utilizes the hidden state of the trained BERT for entity representation, while ProbExpan relies on the probability distribution of candidate entities at the [MASK] token. We believe that the hidden state, as a continuous vector in the feature space, captures the semantics of entities with finer granularity. In contrast, the probability distribution, as a discrete metric in the probability space, inherently offers relatively coarser granularity due to its limited capacity to store entity information. Consequently, ProbExpan underperforms in Ultra-ESE. This insight prompts future research into devising better representations to express the ultra-fine-grained semantics of entities, such as decoupling the base semantics of entities from the ultra-fine-grained attribute semantics, similar to the Mix-of-Expert (MoE) approach, where distinct features represent different perspectives of the semantics.

The ablation experiments presented in Table \ref{tab:basic_ablation} further validate our hypothesis regarding the factors why RetExpan and GenExpan surpass the previous state-of-the-art models. Both methods exhibit remarkable degradation after removing the corresponding modules.

(3) Comparing the two enhancement strategies, ultra-fine-grained contrastive learning provides the most substantial improvement in the Pos class metrics, averaging 4.10 points (from 52.04 to 56.14). Conversely, contrastive learning brings smaller gains in Neg class metrics. This discrepancy is primarily attributed to the implementation of contrastive learning, where both positive and negative target entities are pulled away from normal negative entities originating from other semantic classes to ensure underlying semantics. This dilutes the pulling away of positive and negative target entities from each other, as they are both in the denominator of the contrastive loss and are assigned the same weight. 

Furthermore, we found that directly increasing the weights of negative terms formed by positive and negative target entities is ineffective. This is because positive and negative target entities are determined by GPT-4 and inevitably contain errors. Therefore, this inspires us to devise more precise contrastive data mining methods in the future to amplify the penalty for hard negative terms formed by positive and negative target entities.

(4) The conventional probability-based methods SetExpan and CaSE attain quite low scores on both NegMAP and NegP metrics, which indicates that negative target entities are not excessively introduced. However, this does not imply that these methods exhibit a high degree of negative semantic awareness. The Pos class metrics of SetExpan and CaSE are low in similarity, indicating their limited understanding of fine-grained semantic classes overall. Consequently, they struggle to recall entities with the given seed entities, leading to low scores in both Pos and Neg class metrics. We further evaluated the MAP@100 of CaSE at the fine-grained semantic class level, which only reaches 21.43, significantly lower than other methods such as 82.08 of RetExpan.

(5) Without further fine-tuning and modifications to the model structure, GPT-4 achieves excellent results, which proves that generative LLMs have great potential for addressing Ultra-ESE tasks. However, GPT-4 still fails to outperform the GenExpan, which is also generative but based on smaller LLaMA-7B. Analysis of cases reveals that there are two primary problems with GPT-4. On the one hand, it performs poorly on long-tail problems, such as U.S. national monuments and mobile phone brands, which contain a considerable number of low-frequency entities with limited information available on the Internet. GPT-4 only achieves single-digit PosMAP on these semantic classes. In contrast, GenExpan performs better, benefiting from the given contextual corpus. On the other hand, GPT-4 is prone to haphazardly generate non-existent entities (e.g., fake mobile phone brands), which is referred to hallucination problem in recent work\cite{ji2023towards,cheng2023accelerating}. We are currently unable to solve this issue by simple output post-processing. While the results of GPT-4 are impressive, there remains a lot of space for improvement.

\begin{table}[]
\caption{Ablation experiments for each module of RetExpan and GenExpan. Each module contributes to the overall performance.}
\begin{tabular}{lccccc}
\toprule
\multirow{2}{*}{\textbf{Method}} & \multicolumn{4}{c}{\textbf{MAP}}       & \multirow{2}{*}{\textbf{Avg}} \\ \cmidrule{2-5}
                        & \textbf{@10}   & \textbf{@20}   & \textbf{@50}   & \textbf{@100}  &                      \\ \midrule
RetExpan                & 66.48 & 65.25 & 63.95 & 63.31 & 64.75              \\ \midrule
- Entity prediction     & 63.94 & 62.27 & 61.32 & 60.48 & 62.00              \\ \midrule\midrule
GenExpan                & 69.79 & 68.35 & 67.07 & 66.38 & 67.90              \\ \midrule
- Prefix constrain      & 57.03 & 56.64 & 56.33 & 56.1  & 56.53               \\ \midrule
- Further pretrain      & 68.58 & 66.67 & 65.23 & 64.23 & 66.18             \\ \bottomrule
\end{tabular}
\label{tab:basic_ablation}
\vspace{-0.5cm}
\end{table}

\subsection{Analysis of Negative Entities}
We conducted analytical experiments from various perspectives regarding negative seed entities to answer the following questions:

\noindent\textbf{Whether the introduction of negative seed entities is effective for representing ultra-fine-grained semantic classes?} The negative seed entity-based entity re-ranking modules were removed from RetExpan and GenExpan. Thanks to the high scalability, it was also integrated into ProbExpan. Results of the three comparison experiments are presented in Table \ref{tab:wo_neg_rerank}. We can observe that:

(1) After discarding the input of negative samples, RetExpan and GenExpan show a consistent decrease and increase in Pos class metrics and Neg class metrics, respectively. ProbExpan, on the other hand, demonstrates a rise of 0.33 points on average in the Comb class metrics after equipping the entity re-ranking module. This strongly suggests that the introduction of negative seed entities effectively mitigates the intrusion of negative target entities while raising the ranking of positive target entities, i.e., portraying the semantic classes at a finer granularity.

(2) Further analyzing the Pos and Neg metrics, we observe that the addition of entity re-ranking module to ProbExpan conversely resulted in a notable decline in PosP metrics. It's essential to clarify the distinction between  MAP and P metrics: P@K is concerned solely with number of target entities in the top-K entity list, regardless of their positions, while MAP@K is rank-aware, with higher values indicating that target entities are positioned closer to the top of the list. Consequently, this indicates that the incorporation of negative seed entities in ProbExpan makes a mixed impact, propelling some positive target entities to higher ranks and simultaneously causing others to be ranked lower.

(3) For RetExpan, negative seed entities have a similar impact on metrics with different K values, whereas for GenExpan, negative seed entities have a greater impact on metrics with a smaller K. This may be due to the fact that RetExpan is a one-time expansion~\cite{kushilevitz2020two,mamou2018term,yu2019corpus} model and GenExpan is an iterative expansion~\cite{AuxiliaryExpan,rong2016egoset,shen2017setexpan} model, so entities expanded later (corresponding to a larger K) are more likely to deviate from the original ground-truth semantic class. During these stages, re-ranking using negative seed entities also doesn't work. Addressing the issue of semantic drift~\cite{curran2007minimising,mcintosh2010unsupervised,shi2014probabilistic} in iterative expansion models like GenExpan is a longstanding challenge in the ESE field.

\begin{table*}[th]
\centering
\caption{Ablation experiments on entity re-ranking module with negative seed entities.}
\scalebox{0.85}{
\begin{tabular}{ll|cccc|cccc|c}
\toprule
                                                                            &                                                                                  & \multicolumn{4}{c}{\textbf{MAP}}                                                                                              & \multicolumn{4}{c}{\textbf{P}}                                                                                                &                                \\ \cmidrule{3-10}
\multirow{-2}{*}{\textbf{Method}}                                           & \multirow{-2}{*}{\textbf{\begin{tabular}[c]{@{}l@{}}Metric\\ Type\end{tabular}}} & \textbf{@10}                  & \textbf{@20}                  & \textbf{@50}                  & \textbf{@100}                 & \textbf{@10}                  & \textbf{@20}                  & \textbf{@50}                  & \textbf{@100}                 & \multirow{-2}{*}{\textbf{Avg}} \\ \midrule
                                                                            & Pos                                                                     & 21.86                         & 22.11                         & 22.80                         & 23.89                         & 38.08                         & 39.41                         & 47.02                         & 62.71                         & 34.74                         \\
                                                                            & Neg                                                                     & 6.72                          & 8.16                          & 10.85                         & 13.47                         & 15.12                         & 19.92                         & 34.51                         & 56.48                         & 20.65                         \\
\multirow{-3}{*}{ProbExpan}                                                 & Comb                                                                    & 57.57                         & 56.98                         & 55.97                         & 55.21                         & 61.48                         & 59.75                         & 56.25                         & 53.12                         & 57.04                         \\ \cmidrule{2-11} 
                                                                            & Pos                                                                     & 23.63                         & 23.65                         & 24.13                         & 25.20                         & 37.36                         & 39.41                         & 46.79                         & 62.71                         & 35.36                         \\
                                                                            & Neg                                                                     & 6.82                          & 8.22                          & 10.89                         & 13.47                         & 14.79                         & 19.92                         & 34.48                         & 56.48                         & 20.63                         \\
\multirow{-3}{*}{+ Neg Rerank}                                              & Comb                                                                    & 58.41                         & 57.72                         & 56.62                         & 55.87                         & 61.29                         & 59.75                         & 56.20                         & 53.12                         & 57.37                         \\ \cmidrule{2-11} 
                                                                            & Pos                                                                     & \cellcolor[HTML]{EFEFEF}1.77  & \cellcolor[HTML]{EFEFEF}1.54  & \cellcolor[HTML]{EFEFEF}1.33  & \cellcolor[HTML]{EFEFEF}1.32  & \cellcolor[HTML]{EFEFEF}-0.72 & \cellcolor[HTML]{EFEFEF}0.00  & \cellcolor[HTML]{EFEFEF}-0.23 & \cellcolor[HTML]{EFEFEF}0.00  & \cellcolor[HTML]{EFEFEF}0.63  \\
                                                                            & Neg                                                                     & \cellcolor[HTML]{EFEFEF}0.10  & \cellcolor[HTML]{EFEFEF}0.07  & \cellcolor[HTML]{EFEFEF}0.04  & \cellcolor[HTML]{EFEFEF}0.00  & \cellcolor[HTML]{EFEFEF}-0.33 & \cellcolor[HTML]{EFEFEF}0.00  & \cellcolor[HTML]{EFEFEF}-0.13 & \cellcolor[HTML]{EFEFEF}0.00  & \cellcolor[HTML]{EFEFEF}-0.03 \\
\multirow{-3}{*}{$\triangle$}                                                       & Comb                                                                    & \cellcolor[HTML]{EFEFEF}0.83  & \cellcolor[HTML]{EFEFEF}0.74  & \cellcolor[HTML]{EFEFEF}0.64  & \cellcolor[HTML]{EFEFEF}0.66  & \cellcolor[HTML]{EFEFEF}-0.19 & \cellcolor[HTML]{EFEFEF}0.00  & \cellcolor[HTML]{EFEFEF}-0.05 & \cellcolor[HTML]{EFEFEF}0.00  & \cellcolor[HTML]{EFEFEF}0.33  \\ \midrule
                                                                            & Pos                                                                     & 41.73                         & 39.53                         & 38.55                         & 39.91                         & 54.58                         & 58.03                         & 66.76                         & 77.23                         & 52.04                         \\
                                                                            & Neg                                                                     & 8.77                          & 9.04                          & 10.65                         & 13.29                         & 16.44                         & 21.04                         & 34.78                         & 56.54                         & 21.32                         \\
\multirow{-3}{*}{\begin{tabular}[c]{@{}l@{}}RetExpan\\ (Ours)\end{tabular}} & Comb                                                                    & 66.48                         & 65.25                         & 63.95                         & 63.31                         & 69.07                         & 68.50                         & 65.99                         & 60.34                         & 65.36                         \\ \cmidrule{2-11} 
                                                                            & Pos                                                                     & 40.39                         & 38.33                         & 37.31                         & 38.70                         & 54.09                         & 58.03                         & 66.62                         & 77.23                         & 51.34                         \\
                                                                            & Neg                                                                     & 9.31                          & 9.68                          & 11.44                         & 13.99                         & 16.81                         & 21.04                         & 35.53                         & 56.54                         & 21.79                         \\
\multirow{-3}{*}{- Neg Rerank}                                              & Comb                                                                    & 65.54                         & 64.33                         & 62.94                         & 62.36                         & 68.64                         & 68.50                         & 65.55                         & 60.34                         & 64.78                         \\ \cmidrule{2-11} 
                                                                            & Pos                                                                     & \cellcolor[HTML]{EFEFEF}-1.34 & \cellcolor[HTML]{EFEFEF}-1.20 & \cellcolor[HTML]{EFEFEF}-1.24 & \cellcolor[HTML]{EFEFEF}-1.21 & \cellcolor[HTML]{EFEFEF}-0.49 & \cellcolor[HTML]{EFEFEF}0.00  & \cellcolor[HTML]{EFEFEF}-0.14 & \cellcolor[HTML]{EFEFEF}0.00  & \cellcolor[HTML]{EFEFEF}-0.70 \\
                                                                            & Neg                                                                     & \cellcolor[HTML]{EFEFEF}0.54  & \cellcolor[HTML]{EFEFEF}0.64  & \cellcolor[HTML]{EFEFEF}0.79  & \cellcolor[HTML]{EFEFEF}0.70  & \cellcolor[HTML]{EFEFEF}0.37  & \cellcolor[HTML]{EFEFEF}0.00  & \cellcolor[HTML]{EFEFEF}0.75  & \cellcolor[HTML]{EFEFEF}0.00  & \cellcolor[HTML]{EFEFEF}0.47  \\  
\multirow{-3}{*}{$\triangle$}                                                       & Comb                                                                    & \cellcolor[HTML]{EFEFEF}-0.94 & \cellcolor[HTML]{EFEFEF}-0.92 & \cellcolor[HTML]{EFEFEF}-1.01 & \cellcolor[HTML]{EFEFEF}-0.95 & \cellcolor[HTML]{EFEFEF}-0.43 & \cellcolor[HTML]{EFEFEF}0.00  & \cellcolor[HTML]{EFEFEF}-0.44 & \cellcolor[HTML]{EFEFEF}0.00  & \cellcolor[HTML]{EFEFEF}-0.59 \\ \midrule
                                                                            & Pos                                                                     & 46.79                         & 45.00                         & 42.89                         & 40.80                         & 59.77                         & 62.15                         & 66.26                         & 66.57                         & 53.78                         \\
                                                                            & Neg                                                                     & 7.25                          & 8.28                          & 8.72                          & 8.01                          & 15.21                         & 21.31                         & 27.33                         & 28.58                         & 15.59                         \\
\multirow{-3}{*}{\begin{tabular}[c]{@{}l@{}}GenExpan\\ (Ours)\end{tabular}} & Comb                                                                    & 69.77                         & 68.36                         & 67.08                         & 66.40                         & 72.28                         & 70.42                         & 69.47                         & 68.99                         & 69.10                         \\ \cmidrule{2-11} 
                                                                            & Pos                                                                     & 46.08                         & 44.31                         & 42.17                         & 40.10                         & 59.26                         & 62.15                         & 66.17                         & 66.57                         & 53.35                         \\
                                                                            & Neg                                                                     & 7.85                          & 8.71                          & 9.20                          & 8.45                          & 15.70                         & 21.31                         & 27.44                         & 28.58                         & 15.90                         \\
\multirow{-3}{*}{-Neg Rerank}                                               & Comb                                                                    & 69.12                         & 67.80                         & 66.49                         & 65.83                         & 71.78                         & 70.42                         & 69.37                         & 68.99                         & 68.98                         \\ \cmidrule{2-11} 
                                                                            & Pos                                                                     & \cellcolor[HTML]{EFEFEF}-0.71  & \cellcolor[HTML]{EFEFEF}-0.69  & \cellcolor[HTML]{EFEFEF}-0.72 & \cellcolor[HTML]{EFEFEF}-0.70  & \cellcolor[HTML]{EFEFEF}-0.51 & \cellcolor[HTML]{EFEFEF}0.00 & \cellcolor[HTML]{EFEFEF}-0.09 & \cellcolor[HTML]{EFEFEF}0.00 & \cellcolor[HTML]{EFEFEF}-0.43  \\
                                                                            & Neg                                                                     & \cellcolor[HTML]{EFEFEF}0.60  & \cellcolor[HTML]{EFEFEF}0.43  & \cellcolor[HTML]{EFEFEF}0.48  & \cellcolor[HTML]{EFEFEF}0.44  & \cellcolor[HTML]{EFEFEF}0.49  & \cellcolor[HTML]{EFEFEF}0.00 & \cellcolor[HTML]{EFEFEF}0.11  & \cellcolor[HTML]{EFEFEF}0.00  & \cellcolor[HTML]{EFEFEF}0.32  \\
\multirow{-3}{*}{$\triangle$}                                                       & Comb                                                                    & \cellcolor[HTML]{EFEFEF}-0.65 & \cellcolor[HTML]{EFEFEF}-0.56 & \cellcolor[HTML]{EFEFEF}-0.59 & \cellcolor[HTML]{EFEFEF}-0.57 & \cellcolor[HTML]{EFEFEF}-0.50 & \cellcolor[HTML]{EFEFEF}0.00  & \cellcolor[HTML]{EFEFEF}-0.10 & \cellcolor[HTML]{EFEFEF}0.00  & \cellcolor[HTML]{EFEFEF}-0.37 \\ \bottomrule
\end{tabular}}
\vspace{-0.25cm}
\label{tab:wo_neg_rerank}
\end{table*}

\begin{table}[]
\centering
\caption{Comparison experiments when positive and negative attributes are the same and different.}
\begin{tabular}{llccccc}
\toprule
\multirow{2}{*}{\textbf{Method}}                                                        & \multirow{2}{*}{\begin{tabular}[c]{@{}l@{}}\textbf{Metric}\\ \textbf{Type}\end{tabular}} & \multicolumn{4}{c}{\textbf{MAP}}       & \multirow{2}{*}{\textbf{Avg}} \\ \cmidrule{3-6}
                                                                               &                                                                        & \textbf{@10}   & \textbf{@20}   & \textbf{@50}   & \textbf{@100}  &                      \\ \midrule
\multicolumn{7}{c}{$\mathbf{\mathcal A^{pos}=\mathcal A^{neg}}$}                                                                                                                                                                                  \\ \midrule
\multirow{4}{*}{RetExpan}                                                      & Pos                                                                    & 43.14 & 41.74 & 41.93 & 42.89 & 42.43                \\
                                                                               & Neg                                                                    & 5.54  & 6.44  & 8.71  & 11.56 & 8.06                 \\
                                                                               & Comb                                                                   & 68.80 & 67.65 & 66.61 & 65.67 & 67.18                             \\ \midrule
\multirow{4}{*}{\begin{tabular}[c]{@{}l@{}}RetExpan\\ + Contrast\end{tabular}} & Pos                                                                    & 49.68 & 48.73 & 48.77 & 49.50 & 49.17                \\
                                                                               & Neg                                                                    & 5.76  & 6.51  & 8.64  & 11.62 & 8.13                 \\
                                                                               & Comb                                                                   & 71.96 & 71.11 & 70.06 & 68.94 & 70.52                    \\  \toprule 
\multicolumn{7}{c}{$\mathbf{\mathcal A^{pos}\neq\mathcal A^{neg}}$}                                                                                                                                                                                 \\ \toprule
\multirow{4}{*}{RetExpan}                                                                       & Pos                                                                    & 41.73 & 39.53 & 38.55 & 39.91 & 39.93                \\
                                                                               & Neg                                                                    & 8.77  & 9.04  & 10.65 & 13.29 & 10.44                \\
                                                                               & Comb                                                                   & 66.48 & 65.25 & 63.95 & 63.31 & 64.75                              \\ \midrule
\multirow{4}{*}{\begin{tabular}[c]{@{}l@{}}RetExpan\\ + Contrast\end{tabular}} & Pos                                                                    & 46.44 & 43.69 & 42.51 & 43.14 & 43.95                \\
                                                                               & Neg                                                                    & 8.22  & 9.25  & 11.17 & 13.38 & 10.51                \\
                                                                               & Comb                                                                   & 69.11 & 67.22 & 65.67 & 64.88 & 66.72                             \\  \bottomrule
\end{tabular}
\label{tab:neg_pos_equal}
\end{table}

\begin{table}[th]
\centering
\caption{Comparison experiments on semantic classes with different numbers of positive and negative attributes.}
\renewcommand\arraystretch{1}
\setlength\tabcolsep{2.5pt}
\begin{tabular}{clccccc}
\toprule
\multirow{2}{*}{$(|\mathcal{A}^{pos}|, |\mathcal{A}^{neg}|)$} & \multirow{2}{*}{\textbf{\begin{tabular}[c]{@{}l@{}}Metric\\ Type\end{tabular}}} & \multicolumn{4}{c}{\textbf{MAP}}                           & \multirow{2}{*}{\textbf{Avg}} \\ \cmidrule{3-6}
                                &                                                                                 & \textbf{@10} & \textbf{@20} & \textbf{@50} & \textbf{@100} &                               \\ \midrule
\multirow{3}{*}{(1, 1)}         & Pos                                                                             & 41.32        & 39.18        & 38.36        & 39.58         & 39.61                         \\
                                & Neg                                                                             & 8.61         & 9.14         & 10.88        & 13.43         & 10.52                         \\
                                & Comb                                                                            & 66.36        & 65.02        & 63.74        & 63.07         & 64.55                         \\ \midrule
\multirow{3}{*}{(1, 2)}          & Pos                                                                             & 39.76        & 37.65        & 41.49        & 43.44         & 40.59                         \\
                                & Neg                                                                             & 0.47         & 0.98         & 2.44         & 5.57          & 2.37                          \\
                                & Comb                                                                            & 69.64        & 68.33        & 69.52        & 68.93         & 69.11                         \\ \midrule
\multirow{3}{*}{(2, 1)}          & Pos                                                                             & 38.84        & 32.85        & 36.39        & 41.33         & 37.35                         \\
                                & Neg                                                                             & 2.39         & 3.29         & 5.38         & 9.81          & 5.22                          \\
                                & Comb                                                                            & 68.22        & 64.78        & 65.51        & 65.76         & 66.07                        \\ \bottomrule
\end{tabular}
\label{tab:neg_pos_num}
\end{table}

\noindent\textbf{What are the roles played by negative seed entities?} We first assessed the efficacy of RetExpan and its enhancement strategies by comparing scenarios where positive and negative attributes are identical or different. As shown in Table \ref{tab:neg_pos_equal}, we can find that:

(1) When positive and negative attributes are identical, there is no overlap between positive and negative target entities, which simplifies semantic classes and yielding higher performance. In such cases, negative seed entities play a supportive role, primarily emphasizing user-concerned attributes and mitigating ambiguity. Conversely, when positive and negative attributes are different, the potential overlap between entities satisfying these attributes poses a significant challenge to semantic understanding, leading to lower overall performance. Here, negative seed entities play a crucial role in conveying unwanted semantics and are indispensable.

(2) The benefits derived from contrastive learning are more pronounced in semantic classes where positive and negative attributes are identical. This is because, in such scenarios, there is no overlap between positive and negative target entities, making the contrastive training pairs mined based on the similarity of positive and negative seed entities more reliable.

(3) When positive and negative attributes differ, the primary focus lies in evaluating the model's capability to comprehend the semantics expressed by negative seed entities. We observe that contrastive learning demonstrates less effective in this situation. This phenomenon suggests that the understanding of negative seed entities demands more additional knowledge compared to positive seed classes. This notion is further verified in the analysis of chain-of-thought reasoning in Section \ref{sec:enhance_stra}.

\noindent\textbf{What are the differences in challenges posed by semantic classes with varying numbers of positive and negative attributes?} We analyze the performance of RetExpan on semantic classes characterized by varying quantities of positive and negative attributes in Table \ref{tab:neg_pos_num}. To maintain control over variables, we exclusively present semantic classes for three combinations of attribute quantity: (1,1), (1,2), and (2,1). Imposing stricter constraints on the attributes results in a reduction in the number of associated target entities, rendering it more challenging to discern these entities accurately. Consequently, a decrease in the Pos (Neg) metrics is observed, correlating with an increase in the number of positive (negative) attributes. It's important to emphasize that although the Neg metrics decrease with more negative attribute constraints, this does not imply a reduced difficulty in distinguishing negative target entities. Rather, identifying these negative target entities requires more comprehensive attribute information about negative seed entities.

\subsection{Analysis of Enhancement Strategy}
\label{sec:enhance_stra}
\begin{table}[]
\centering
\caption{Ablation experiments on contrastive learning.}
\renewcommand\arraystretch{1}
\setlength\tabcolsep{2.3pt}
\begin{tabular}{llccccc}
\toprule
\multirow{2}{*}{\textbf{Method}}                                                            & \multirow{2}{*}{\textbf{\begin{tabular}[c]{@{}l@{}}Metric\\ Type\end{tabular}}} & \multicolumn{4}{c}{\textbf{MAP}}                           & \multirow{2}{*}{\textbf{Avg}} \\ \cmidrule{3-6}
                                                                                            &                                                                                 & \textbf{@10} & \textbf{@20} & \textbf{@50} & \textbf{@100} &  \\ \midrule
                                                                                            \multirow{3}{*}{RetExpan}                                                                      & Pos                                                                             & 41.73        & 39.53        & 38.55        & 39.91         & 39.93                         \\
                                                                                               & Neg                                                                             & 8.77         & 9.04         & 10.65        & 13.29         & 10.43                         \\
                                                                                               & Comb                                                                            & 66.48        & 65.25        & 63.95        & 63.31         & 64.75                         \\ \midrule\midrule
\multirow{3}{*}{\begin{tabular}[c]{@{}l@{}}RetExpan\\ +Contrast\end{tabular}}                                                                        & Pos                                                                    & 47.45  & 44.68  & 43.63   & 44.20   & 44.99                \\
                                                                                            & Neg                                                                    & 8.02   & 8.98   & 10.89   & 13.05   & 10.24                \\
                                                                                            & Comb                                                                   & 69.72  & 67.85  & 66.37   & 65.57   & 67.38                \\ \midrule
\multirow{3}{*}{\begin{tabular}[c]{@{}l@{}}- Neg from\\  ($L_{pos}$, $L_{neg}$)\end{tabular}}             & Pos                                                                    & 46.34  & 43.60  & 42.69   & 43.36   & 43.99                \\
                                                                                            & Neg                                                                    & 8.71   & 9.23   & 11.16   & 13.30   & 10.60                \\
                                                                                        
                                                                                            & Comb                                                                   & 68.82  & 67.18  & 65.76   & 65.03   & 66.70                \\ \midrule
\multirow{3}{*}{\begin{tabular}[c]{@{}l@{}}- Neg from \\ ($L_{pos}$, $\overline{L_0}$) \\ \& ($L_{neg}$, $\overline{L_0}$)\end{tabular}} & Pos                                                                    & 46.89  & 44.14  & 43.04   & 43.86   & 44.48                \\
                                                                                            & Neg                                                                    & 8.42   & 9.00   & 11.14   & 13.26   & 10.46                \\
                                                                                         
                                                                                            & Comb                                                                   & 69.24  & 67.57  & 65.95   & 65.30   & 67.01                \\ \midrule

\multirow{3}{*}{\begin{tabular}[c]{@{}l@{}}- Pos from \\ ($L_{pos}$, $L_{pos}$) \\ \& ($L_{neg}$, $L_{neg}$)\end{tabular}}             & Pos                                                                    & 46.95  & 43.55  & 42.77   & 43.36   & 44.16                \\
                                                                                            & Neg                                                                    & 8.531  & 9.103  & 10.996  & 13.185  & 10.45                \\
                                                                                            & Comb                                                                   & 69.211 & 67.221 & 65.8875 & 65.0875 & 66.85                \\ \bottomrule
\end{tabular}
\label{tab:com_cotrastive}
\end{table}

\subsubsection{Analysis of Contrastive Learning}
We conducted ablation experiments on each part of the training data for ultra-fine-grained contrastive learning to explore their impact on overall performance. As illustrated in Table \ref{tab:com_cotrastive}, based on the full training data, the last three rows sequentially remove the hard negative samples comprising pairs of positive and negative target entities, the normal negative samples consisting of pairs of positive and negative target entities along with entities from other semantic classes, and the positive samples composed of entity pairs within the same ultra-fine-grained semantic class.

It can be seen that hard negative samples consisting of pairs of positive and negative target entities contribute the most to the overall performance, which is in line with our intuition of proposing ultra-fine-grained contrastive learning. We set both $|L_{pos}|$ and $|L_{neg}|$ to 10 to mitigate the introduction of excessive noise. Consequently, it's evident that this part of training data wields more influence when K is small. Exploring how to mitigate the noise introduced by automatic GPT-4 labeling when $|L_{pos}|$ and $|L_{neg}|$ increase is a promising direction for future exploration. 

Additionally, we observe that normal negative samples boost the overall performance. They ensure that positive and negative target samples do not deviate excessively from each other at the fine-grained semantic level, thereby preventing semantic collapse. Nevertheless, the integration of normal negative samples compromises the efficacy of ultra-fine-grained (hard) negative samples. This arises from the fact that the presence of normal negative samples indirectly diminishes the penalization applied to the distance of hard negative samples by the loss function, as they are all in the denominator of the loss function and have the same weights. Specifically, the incorporation of normal negative samples diminishes the average enhancement of hard negative samples on the Comb metric from 67.01 - 64.75 = 2.26 (line 4 minus line 1) to 67.38 - 66.70 = 0.68 (line 2 minus line 3). Moreover, positive samples, composed of entities belonging to the same ultra-fine-grained semantic class, similarly contribute positively to the overall performance.

\subsubsection{Analysis of Chain-of-thought Reasoning}
\begin{table}[]
\centering
\caption{Comparison experiments on chain-of-thought reasoning. The impact of reasoning depth and reasoning precision on performance is investigated. GT: Ground-truth. Gen: Generated. CN: Class name. Pos/Neg: Positive/Negative attribute information.}
\renewcommand\arraystretch{1}
\setlength\tabcolsep{2.5pt}
\scalebox{0.85}{
\begin{tabular}{llccccc}
\toprule
\multirow{2}{*}{\textbf{Method}}                                                                             & \multirow{2}{*}{\textbf{\begin{tabular}[c]{@{}l@{}}Metric\\ Type\end{tabular}}} & \multicolumn{4}{c}{\textbf{MAP}}                           & \multirow{2}{*}{\textbf{Avg}} \\ \cmidrule{3-6}
                                                                                                             &                                                                                 & \textbf{@10} & \textbf{@20} & \textbf{@50} & \textbf{@100} &                               \\ \midrule
\multirow{3}{*}{GenExpan}                                                                                    & Pos                                                                             & 46.84        & 44.99        & 42.88        & 40.79         & 43.88                         \\
                                                                                                             & Neg                                                                             & 7.27         & 8.29         & 8.74         & 8.02          & 8.08                          \\
                                                                                                             & Comb                                                                            & 69.79        & 68.35        & 67.07        & 66.38         & 67.90                         \\ \midrule\midrule
\multirow{3}{*}{\begin{tabular}[c]{@{}l@{}}GenExpan\\ + CoT (GT CN)\end{tabular}}                            & Pos                                                                             & 49.20        & 46.84        & 43.67        & 41.04         & 45.19                         \\
                                                                                                             & Neg                                                                             & 7.43         & 9.15         & 9.17         & 8.36          & 8.53                          \\
                                                                                                             & Comb                                                                            & 70.89        & 68.84        & 67.25        & 66.34         & 68.33                         \\ \midrule
\multirow{3}{*}{\begin{tabular}[c]{@{}l@{}}GenExpan\\ + CoT (Gen CN)\end{tabular}}                           & Pos                                                                             & 49.33        & 46.97        & 43.51        & 40.82         & 45.15                         \\
                                                                                                             & Neg                                                                             & 7.37         & 8.97         & 8.74         & 7.89          & 8.24                          \\
                                                                                                             & Comb                                                                            & 70.98        & 69.00        & 67.38        & 66.46         & 68.46                         \\ \midrule\midrule
\multirow{3}{*}{\begin{tabular}[c]{@{}l@{}}GenExpan\\ + CoT (Cen CN \\ \& Gen Pos)\end{tabular}}              & Pos                                                                             & 50.39        & 47.80        & 43.67        & 40.06         & 45.48                         \\
                                                                                                             & Neg                                                                             & 7.79         & 9.29         & 8.15         & 6.89          & 8.03                          \\
                                                                                                             & Comb                                                                            & 71.30        & 69.25        & 67.76        & 66.58         & 68.72                         \\ \midrule
\multirow{3}{*}{\begin{tabular}[c]{@{}l@{}}GenExpan\\ + CoT (Cen CN \\ \& GT Pos)\end{tabular}}               & Pos                                                                             & 50.77        & 48.25        & 44.38        & 40.67         & 46.02                         \\
                                                                                                             & Neg                                                                             & 7.68         & 9.16         & 8.12         & 6.83          & 7.95                          \\
                                                                                                             & Comb                                                                            & 71.55        & 69.54        & 68.13        & 66.92         & 69.04                         \\ \midrule\midrule
\multirow{3}{*}{\begin{tabular}[c]{@{}l@{}}GenExpan\\ + CoT (Cen CN \\ \& Gen Pos \& Gen Neg)\end{tabular}} & Pos                                                                             & 49.78        & 47.24        & 42.68        & 39.01         & 44.68                         \\
                                                                                                             & Neg                                                                             & 7.60         & 8.97         & 7.98         & 6.75          & 7.82                          \\
                                                                                                             & Comb                                                                            & 71.09        & 69.14        & 67.35        & 66.13         & 68.43                         \\ \midrule
\multirow{3}{*}{\begin{tabular}[c]{@{}l@{}}GenExpan\\ + CoT (Cen CN \\ \& GT Pos \& GT Neg)\end{tabular}}   & Pos                                                                             & 51.02        & 49.19        & 44.86        & 41.40         & 46.62                         \\
                                                                                                             & Neg                                                                             & 7.30         & 8.70         & 7.59         & 6.53          & 7.53                          \\
                                                                                                             & Comb                                                                            & 71.86        & 70.25        & 68.63        & 67.43         & 69.54                        \\ \bottomrule
\end{tabular}}
\label{tab:cot_ablation}
\end{table}

Experiments in Table \ref{tab:cot_ablation} examine the chain-of-thought with varying reasoning depths and assess the impact of reasoning precision on model performance. From the results, we observe that:
(1) Deeper reasoning proves to be more advantageous for Ultra-ESE, as it facilitates a more explicit understanding of the implicit ultra-fine-grained semantics of seed entities.

(2) Surprisingly, the use of manually labeled ground-truth class names is not superior to those derived through reasoning with the LLaMA-7B. The case study of the generated class names reveals that LLaMA-7B can produce class names encapsulating positive attribute information based on the given positive seed entities. For instance, when given entities about U.S. airports with the ``location'' attribute, GenExpan accurately infers more specific semantic classes that reflect positive attribute information, e.g., ``Airports in Michigan''. This suggests that reasoning results are not necessarily inferior to manually annotated ``ground-truth'' labels, as chain-of-thought reasoning induces LLMs to attend to details that human ignores.

(3) Incorporating negative attribute information obtained from LLaMA-7B's reasoning leads to a decline in model performance. Reasoning about negative attributes poses greater challenges than positive attributes due to constraints from two aspects: first, the negative seed entities are identical on negative attributes; second, the positive seed entities pose different values on negative attributes compared to the negative seed entities. Our analysis reveals that the current LLaMA-7B model struggles to deduce the relevant negative attributes by comparing positive and negative seed entities. This difficulty is particularly pronounced for long-tailed entities (such as ancient musical instruments) and semantic classes with diverse attributes, which indicates that enhancing the ultra-fine-grained perception of entities within LLMs is a promising avenue for future research.

\begin{figure}[t]
    \centering
    \includegraphics[width=0.98\linewidth]{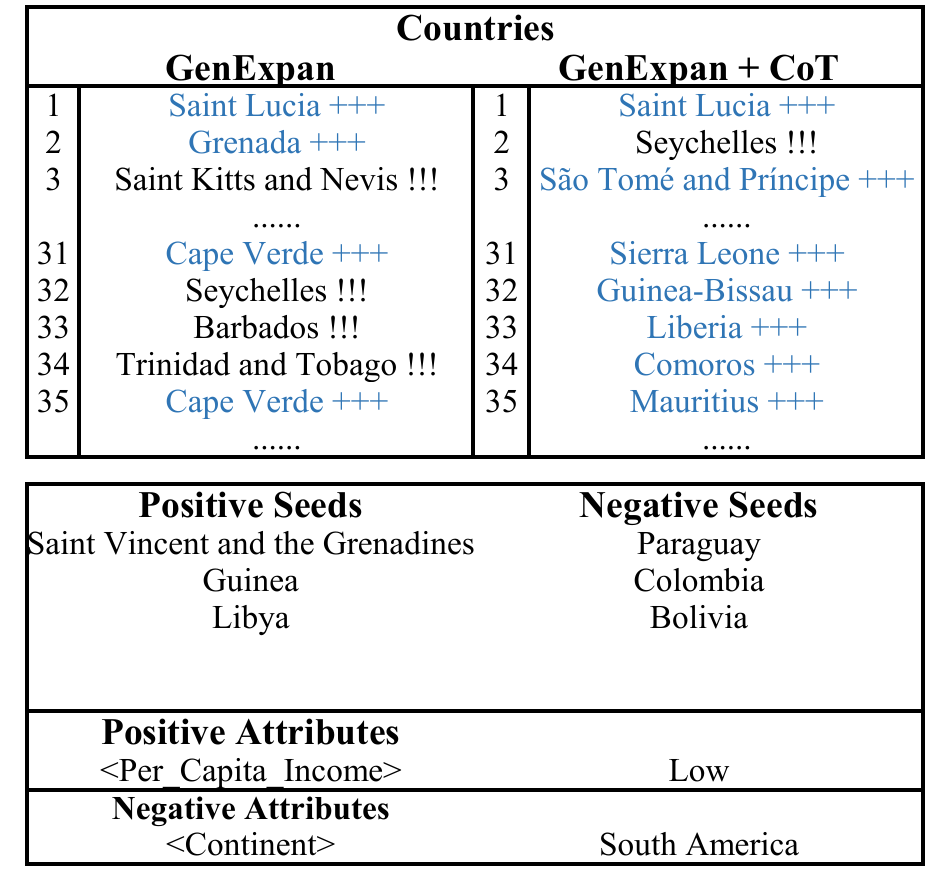}
    \caption{
\textcolor{black}{Case studies of GenExpan and the chain-of-thought reasoning technique. +++ and - - - stand for positive and negative target entities, respectively. !!! represents irrelevant entities belonging to the same fine-grained semantic class as the seed entities.}
    }
    \label{fig:case}
\end{figure}
\subsection{Case Study}
Figure \ref{fig:case} presents some case studies of GenExpan and the chain-of-thought reasoning technique. Overall, it is evident that GenExpan predominantly avoids expanding entities that do not belong to the same semantic class as the positive and negative seed entities. This emphasizes the relative tractability of traditional ESE tasks compared to the significant challenges posed by Ultra-ESE for existing LLMs, where the model cannot exclude the expansion of negative target entities and irrelevant entities (belonging to the same fine-grained semantic class).

For chain-of-thought reasoning, an interesting example is shown in Figure \ref{fig:case}. We find that the ground-truth positive target semantics is ``low-income countries'', but the model erroneously infers the positive semantic class as ``African countries'', which encompasses relatively more low-income countries, and thus improves the recall of the positive target entities. Of course, some high-income African countries such as ``Seychelles'' are also incorrectly introduced. In fact, this error could have been avoided since the provided positive seed entity contains the non-African country ``Saint Vincent and the Grenadines''. This also shows the potential for improvement in the chain-of-thought reasoning strategy.

\section{Conclusion}
In conclusion, our study addresses the challenge of ultra-fine-grained Entity Set Expansion (Ultra-ESE) by introducing negative seed entities alongside positive ones, mitigating ambiguity and facilitating the expression of ``unwanted'' attributes. The creation of UltraWiki, a tailored dataset for Ultra-ESE, enables rigorous evaluation and future research. Through RetExpan and GenExpan frameworks, we demonstrate the effectiveness of large language models in Ultra-ESE from retrieval-based and generation-based perspectives. Our proposed strategies enhance models' comprehension of ultra-fine-grained entity semantics. While our findings are promising, they also highlight the substantial scope for improvement in Ultra-ESE, advancing the field and paving the way for future exploration.

\section*{Acknowledgement}
This research project is supported by National Natural Science Foundation of China (Grant No.62276154), Research Center for ComputerNetwork (Shenzhen) Ministry of Education, the Natural Science Foundation of Guangdong Province (Grant No.2023A1515012914 and 440300241033100801770), Basic Research Fund of Shenzhen City (Grant No.JCYJ20210324120012033, JCYJ20240813112009013 and GJHZ20240218113603006), the Major Key Projectof PCL for Experiments and Applications (PCL2022A05 and PCL2023A09), NSF under Grants III-2106758, and POSE-2346158.

Xuming Hu was supported by the Guangdong Provincial Department of Education Project (Grant No.2024KQNCX028); CAAI-Ant Group Research Fund; Scientific Research Projects for the Higher-educational Institutions (Grant No.2024312096), Education Bureau of Guangzhou Municipality; Guangzhou-HKUST(GZ) Joint Funding Program (Grant No.2025A03J3957), Education Bureau of Guangzhou Municipality.

\bibliographystyle{IEEEtran}
\bibliography{sample}


\end{document}